\documentclass[paper,dvipdfmx]{ieice}
\usepackage{cite}
\usepackage{amsmath,amssymb,amsfonts}
\usepackage{algorithmic}
\usepackage{graphicx}

\usepackage{wrapfig}
\usepackage{xcolor}
\usepackage{color}
\usepackage{multirow}
\usepackage{url}
\usepackage{subfig}
\usepackage{float}
\usepackage{etoolbox}
\usepackage{algorithm}

\def\BibTeX{{\rm B\kern-.05em{\sc i\kern-.025em b}\kern-.08em
		T\kern-.1667em\lower.7ex\hbox{E}\kern-.125emX}}

\usepackage{latexsym}

\usepackage{newtxtext}
\usepackage[varg]{newtxmath}

\field{B}

\title{Feature-based model selection for object detection from point cloud data}

\authorlist{%
\breakauthorline{4}
\authorentry[ktokuda@icn.cce.i.kyoto-u.ac.jp]{Kairi Tokuda}{n}{labelA}\MembershipNumber{}
\authorentry{Ryoichi Shinkuma}{f}{labelB}\MembershipNumber{0001040}
\authorentry{Takehiro Sato}{m}{labelA}\MembershipNumber{1010972}
\authorentry{Eiji Oki}{f}{labelA}\MembershipNumber{9414827}
}

\affiliate[labelA]{The authors are with the
\EICdepartment{\texttt{Graduate School of Informatics}}
\EICorganization{\texttt{Kyoto University}}
\EICaddress{\texttt{Yoshida-honmachi, Sakyo-ku, Kyoto, 606-8501 JAPAN}}
}

\affiliate[labelB]{The author is with the
\EICdepartment{\texttt{Faculty of Engineering}}
\EICorganization{\texttt{Shibaura Institute of Technology}}
\EICaddress{\texttt{Shibaura, Minato-ku, Tokyo 135-8548 JAPAN}}
}

\begin{document}

\maketitle

\begin{summary}
    Smart monitoring using three-dimensional (3D) image sensors has been attracting attention in the context of smart cities. In smart monitoring, object detection from point cloud data acquired by 3D image sensors is implemented for detecting moving objects such as vehicles and pedestrians to ensure safety on the road. However, the features of point cloud data are diversified due to the characteristics of light detection and ranging (LIDAR) units used as 3D image sensors or the install position of the 3D image sensors. Although a variety of deep learning (DL) models for object detection from point cloud data have been studied to date, no research has considered how to use multiple DL models in accordance with the features of the point cloud data. In this work, we propose a feature-based model selection framework that creates various DL models by using multiple DL methods and by utilizing training data with pseudo incompleteness generated by two artificial techniques: sampling and noise adding. It selects the most suitable DL model for the object detection task in accordance with the features of the point cloud data acquired in the real environment. To demonstrate the effectiveness of the proposed framework, we compare the performance of multiple DL models using benchmark datasets created from the KITTI dataset and present example results of object detection obtained through a real outdoor experiment. Depending on the situation, the detection accuracy varies up to 32\% between DL models, which confirms the importance of selecting an appropriate DL model according to the situation.
\end{summary}

\section{Introduction} 
\label{sec:Introduction}
Smart cities aim to provide services that improve the quality of our lives by combining various types of sensors, networks, and data processing technologies. Du \textit{et al.} suggested that smart monitoring systems for smart cities will improve our living conditions in terms of comfort and safety \cite{du2018sensable}. In particular, a smart monitoring system that utilizes information acquired by three-dimensional (3D) image sensors such as a light detection and ranging (LIDAR) unit is expected to be an effective countermeasure against the increasing number of traffic accidents, which have become a problem in recent years \cite{sato2020prioritized}. 

Object detection is a task in computer vision that automatically detects objects as a rectangle or cuboid bounding box by using the data representing real space acquired by the image sensor as input. In recent years, object detection has been attracting attention in various fields, including robotics \cite{mertz2013moving}. Object detection is also required from the perspective of smart monitoring, since it is essential to identify the position of each moving object (e.g., vehicles and pedestrians) in order to ensure safety on the road. The detection accuracy of moving objects impacts the performance of subsequent applications such as multi-object tracking \cite{weng20203d}.

The features of point cloud data acquired in a real environment are diversified due to different measurement situations. 
For example, in situations where relatively inexpensive sensors are used, or where sensors are installed at high altitudes, the density of points drops. This leads to a degradation of data quality. While various deep learning (DL) methods\footnote{In this paper, we refer to specific DL algorithms as DL methods, and inference models trained by the DL methods based on training data as DL models.} for object detection have been developed and have shown competitive performances to date \cite{yan2018second} \cite{lang2019pointpillars} \cite{shi2019pointrcnn} \cite{shi2020points} \cite{shi2020pv}, the performances have typically been measured using datasets that consist of good quality data. Although several studies have provided simple comparisons of DL methods for object detection \cite{guo2020deep} \cite{grigorescu2020survey}, they do not consider data with incompleteness, such as low density of points and noise added to points. In terms of the no free lunch (NFL) theorem \cite{wolpert1995no}, it is difficult to imagine a DL model that would work well for all of the diverse data.

There are several works that consider rules for selecting the learning algorithms that handle simple tasks (such as classification) depending on the features of problems \cite{ali2006learning} \cite{kotthoff2012evaluation}. However, no studies have addressed the selection of DL models for object detection. The algorithms of DL methods for object detection have a broad range of features, and each method can be classified according to which features it has. We need to clarify how the detection accuracies of DL models change when data acquired under adverse conditions are used and how to determine a suitable DL model for each situation accordingly in the field of object detection. 

This paper proposes a feature-based model selection framework for object detection from point cloud data. The proposed framework creates various DL models using multiple DL methods with different features and then selects a model of the method with the most suitable features in accordance with the features of the point cloud data. In order to improve the detection accuracy for data with specific incompleteness, the proposed framework uses training data with pseudo incompleteness generated by two artificial techniques: sampling and noise adding. We report our comparison of the performance of multiple DL models using various data simulating measurement situations created from the KITTI dataset \cite{geiger2012we}. In addition, we present example results of object detection obtained through a real outdoor experiment. Our findings show that even simple features describing the algorithm briefly are capable of explaining the strengths and weaknesses of each model for each situation, which indicates that the concept of the feature-based model selection in the proposed framework is effective.

The main contributions of this paper can be summarized as follows.

\begin{itemize}
	\item We discuss the possible incompleteness of point cloud data and the features of detected objects. Then, we suggest that a DL method that provides high-accuracy object detection differs depending on types of objects and point cloud data, and that we need to use different DL methods according to the situation.
	
	\item We propose a feature-based model selection framework for object detection from point cloud data. In the proposed framework, multiple DL models are created to deal with various situations. The proposed framework selects an appropriate DL method based on the size of the target object and the data quality, and then selects a DL model using training data whose degree of incompleteness is the closest to the inference data. In the proposed framework, training data with pseudo incompleteness is used to improve the detection accuracy of DL models for data with incompleteness.
	
	\item The performance of the proposed framework is evaluated using the KITTI dataset and point cloud data acquired through outdoor experiments using actual equipment. These results confirm the necessity of using different DL methods according to the situation, the effect of adding pseudo incompleteness to the training data, and the effectiveness of the proposed framework.
\end{itemize}

The remainder of this paper is as follows. In Section~\ref{sec:Related work}, we review various DL methods for object detection from point cloud data, approaches to improve the performance of object detection using multiple DL methods, and data augmentation techniques. Section~\ref{sec:Proposed framework} presents the details of the proposed framework. In Section~\ref{sec:Evaluation}, we report the performance evaluations using the KITTI dataset. In Section~\ref{sec:Outdoor}, we discuss examples from an outdoor experiment. We conclude in Section~\ref{sec:Conclusion} with a brief summary.

\section{Related work} 
\label{sec:Related work}
\subsection{Object detection from point cloud data} 
\label{sec:Object detection}
To date, DL methods have been developed extensively to perform object detection. DL methods that use point cloud data acquired by 3D image sensors have been attracting attention recently, since point clouds have rich 3D information. There are also methods that use point clouds and 2D images as input simultaneously\cite{chen2017multi}\cite{qi2018frustum}. This paper assumes the situation where 2D images are not available and focuses on DL methods that use only point cloud data.

A wide variety of DL methods for object detection from point cloud data exists. Some of the algorithms share certain broad features, which can be used to roughly classify the methods into those with common features. In the following, we explain these features, referring to related DL methods. We summarize the DL methods in Table~\ref{tab:Related}.

\subsubsection{Number of stages}
DL methods can be broadly divided into two-stage and one-stage methods depending on the number of stages in the architecture. In the two-stage methods, possible regions containing objects (generally called proposals) are generated on the basis of encoded features of input point clouds in the first stage. Then, in the second stage, the proposals are refined to generate the final bounding boxes. Shi \textit{et al.} suggested that refining the 3D proposals in the second stage improves the detection performance \cite{shi2020points}. In contrast, the one-stage methods estimate the bounding boxes directly. The one-stage methods are generally faster than the two-stage methods due to its simpler structure \cite{lin2017focal}. For this reason, the one-stage methods are often used to pursue real-time performance for applications, such as autonomous driving \cite{yan2018second} \cite{lang2019pointpillars}.

\subsubsection{Processing unit of point clouds}
In 2018, Zhou \textit{et al.} developed VoxelNet \cite{zhou2018voxelnet}. In a voxel-based process such as that used in VoxelNet, irregular point clouds are divided into units of 3D rectangular spaces, which are called voxels, and the features of the points in each voxel are generally aggregated and encoded. Then, a 3D convolutional neural network (CNN) is used to aggregate the voxel-wise features. Part-A$\mathbf{^2} $ net simply takes the average of the features of included points as the initial voxel-wise feature in the first stage \cite{shi2020points}. The voxel-based process is generally computationally efficient due to the regular arrangement of voxels. Submanifold sparse convolution \cite{graham2017submanifold} was adopted to further speed up the process \cite{yan2018second} \cite{shi2020points} \cite{wangcenternet}.

Lang \textit{et al.} claimed that 3D convolution creates a bottleneck in the processing speed and developed PointPillars using 2D convolution as an alternative \cite{lang2019pointpillars}. PointPillars divides point clouds into sets of vertical pillars and then encodes the features of each pillar using the features of the included points. BirdNet also divides the x-y plane into cells and encodes the features of each cell using the  features of included points, such as the number of points~\cite{beltranbirdnet}. The process of handling point clouds for each grid on the x-y plane is called the pillar-based process. 

In contrast, the process of directly handling the raw point cloud without quantization is generally called the point-based process. After Qi \textit{et al.} developed PointNet to directly manipulate raw point cloud data \cite{qi2017pointnet}, point-based methods that use it and its variant, PointNet++, was developed \cite{qi2017pointnet++}. In 2019, Shi \textit{et al.} developed PointRCNN, which performs a point-based process using PointNet++ in all stages to avoid information loss due to quantization \cite{shi2019pointrcnn}. 

Some of the two-stage methods use different processing units for the first and second stages. PV-RCNN performs a voxel-based process similar to Part-A$\mathbf{^2} $ net in the first stage and then a point-based process using PointNet-based networks in the second stage \cite{shi2020pv}. In contrast, STD performs a point-based process using PointNet++ in the first stage and then a voxel-based process in the second stage \cite{yang2019std}. On the other hand, P2V-RCNN adopts a point-to-voxel feature learning approach to improve detection performance, which takes advantage of both structured voxel-based and accurate point-based point cloud representations in the first stage \cite{li2021p2vrcnn}.

\subsubsection{Box generation strategy}
There are two main ways to generate proposals or bounding boxes: an anchor-based strategy and an anchor-free strategy \cite{zhang2020bridging}.

The anchor-based strategy can be generally divided into one-stage methods and two-stage methods. Both of them first tile a large number of preset anchors on the image, then predict the category and refine the coordinates of these anchors by one or several times, and finally output these refined anchors as detection results. Since two-stage methods refine anchors more times than one-stage methods do, the former ones have more accurate results while the latter ones have higher computational efficiency.
In the anchor-based strategy, the region proposal network (RPN) head is adapted to the 2D feature map encoded by the previous convolution layer, and boxes are generated by using anchors for each object class that have been defined in advance \cite{ren2016faster}. To date, several methods including SECOND have used a single shot multibox detector (SSD)-like \cite{liu2016ssd} architecture to build the RPN \cite{yan2018second}. 

In contrast, the anchor-free strategy does not use an anchor box but rather generates boxes directly from foreground points in a bottom-up manner. Shi \textit{et al.} suggested that this strategy is generally lightweight and memory efficient \cite{shi2020points}. The anchor-free strategy directly finds objects without preset anchors in two different ways \cite{zhang2020bridging}. One way is to first locate several pre-defined or self-learned keypoints and then bound the spatial extent of objects. This type of anchor-free strategies is called keypoint-based methods. Another way is to use the center point or region of objects to define positives and then predict the four distances from positives to the object boundary. This kind of anchor-free strategies is called center-based methods. CenterNet3D achieves accurate bounding box regression using the corner attention module, which forces the CNN backbone to pay attention to the object boundaries for effective corner heatmap learning \cite{wangcenternet}. The anchor-free strategy is able to eliminate its hyper-parameters related to anchors and has achieved similar performance with the anchor-based strategy, making anchors more potential in terms of generalization ability.

\begin{table*}[t]
	\caption{Summary of studies on object detection by DL using only point cloud data.}
	\begin{center}
		\renewcommand{\arraystretch}{1.4}
%		\scalebox{0.92}{
			\begin{tabular}{|c|c||c||c|c|c||c|c|}
				\hline
				\textbf{References}&\textbf{DL method}&\textbf{Number of stages}& \textbf{Point-based} & \textbf{Voxel-based} & \textbf{Pillar-based} & \textbf{Anchor-free} & \textbf{Anchor-based}  \\
				\hline
				Yan \textit{et al.} \cite{yan2018second}&SECOND & 1 &  & \checkmark &  &  & \checkmark\\
				\hline
				Lang \textit{et al.} \cite{lang2019pointpillars}&PointPillars & 1 &  &  & \checkmark &  & \checkmark \\
				\hline
				Shi \textit{et al.} \cite{shi2019pointrcnn} &PointRCNN & 2 & \checkmark &  &  & \checkmark &  \\
				\hline
				Shi \textit{et al.} \cite{shi2020points}&Part-A$ ^2 $ & 2 & & \checkmark & & \checkmark & \checkmark \\
				\hline
				Shi \textit{et al.} \cite{shi2020pv}&PV-RCNN & 2 & \checkmark & \checkmark &  &  & \checkmark \\
				\hline
				Zhou \textit{et al.} \cite{zhou2018voxelnet}& VoxelNet & 1 &  & \checkmark &  &  & \checkmark \\
				\hline
				Wang \textit{et al.} \cite{wangcenternet}& CenterNet3D & 1 &  & \checkmark &  & \checkmark &  \\
				\hline
				Beltr\'{a}n \textit{et al.} \cite{beltranbirdnet}&BirdNet & 2 &  &  & \checkmark &  & \checkmark \\
				\hline
				Yang \textit{et al.} \cite{yang2019std}& STD & 2 & \checkmark & \checkmark &  &  & \checkmark \\
				\hline
				Li \textit{et al.} \cite{li2021p2vrcnn}& P2V-RCNN & 2 & \checkmark & \checkmark &  &  & \checkmark \\
				\hline
			\end{tabular}
%		}
		\label{tab:Related}
	\end{center}
\end{table*}

\subsection{Ensemble learning}
Ensemble learning is an approach that aims to improve performance by using multiple machine learning methods, which is said to enhance the generalizability \cite{zhou2009ensemble}. In the field of object detection, box ensemble techniques simply mix the output boxes of each model so that there is only one box per object. Non-maximum suppression (NMS) and soft-NMS are already well used as a way to keep only the most likely correct box from among multiple boxes representing the same object  \cite{bodla2017soft}. Solovyev \textit{et al.} developed Weighted Boxes Fusion, in which the output boxes of each model are weighted and fused in accordance with their scores \cite{solovyev2019weighted}. Casado \textit{et al.} claimed that box mixing blindly risks many false positives, and devised the box ensemble method which uses different voting strategies and is independent of the detector's algorithm \cite{casadoensemble}. Since these ensemble methods need to obtain results from multiple models, the computation cost increases with the number of models used. However, due to the limited computational resources of most edge devices, there is a limit to the number of models that can be handled at the same time, which makes it difficult to take full advantage of the ensemble learning. In our framework, the edge device only uses one model, so only the minimum number of resources is required.

\subsection{Data augmentation}
The technique of adding modified data to the training data is called data augmentation. Shorten \textit{et al.} summarized the data augmentation for DL as a way to avoid overfitting and to train models with high generalizability \cite{shorten2019survey}. However, the more training data are added, the more training time and additional memory are required. There are several data augmentation techniques that add some degraded data to the training data for solving the bias of training data \cite{zhong2020random} \cite{moreno2018forward} \cite{mapointdrop}. Ma \textit{et al.} developed an adversarial data augmentation method, PointDrop, which jointly optimizes the data augmenter and detector to improve robustness for point cloud data with a small number of points \cite{mapointdrop}. Thus, the use of degraded data as training data is common in the context of data augmentation. Differently from such data augmentation, our framework uses only data with pseudo incompleteness in order to deal with the situation where the inference data consists only of data with incompleteness, such as low density of points and noise added to points.

\section{Proposed framework} 
\label{sec:Proposed framework}
\subsection{System model} 
\label{sec:System model}
\begin{figure}[t]
	\centerline{\includegraphics[width=\linewidth]{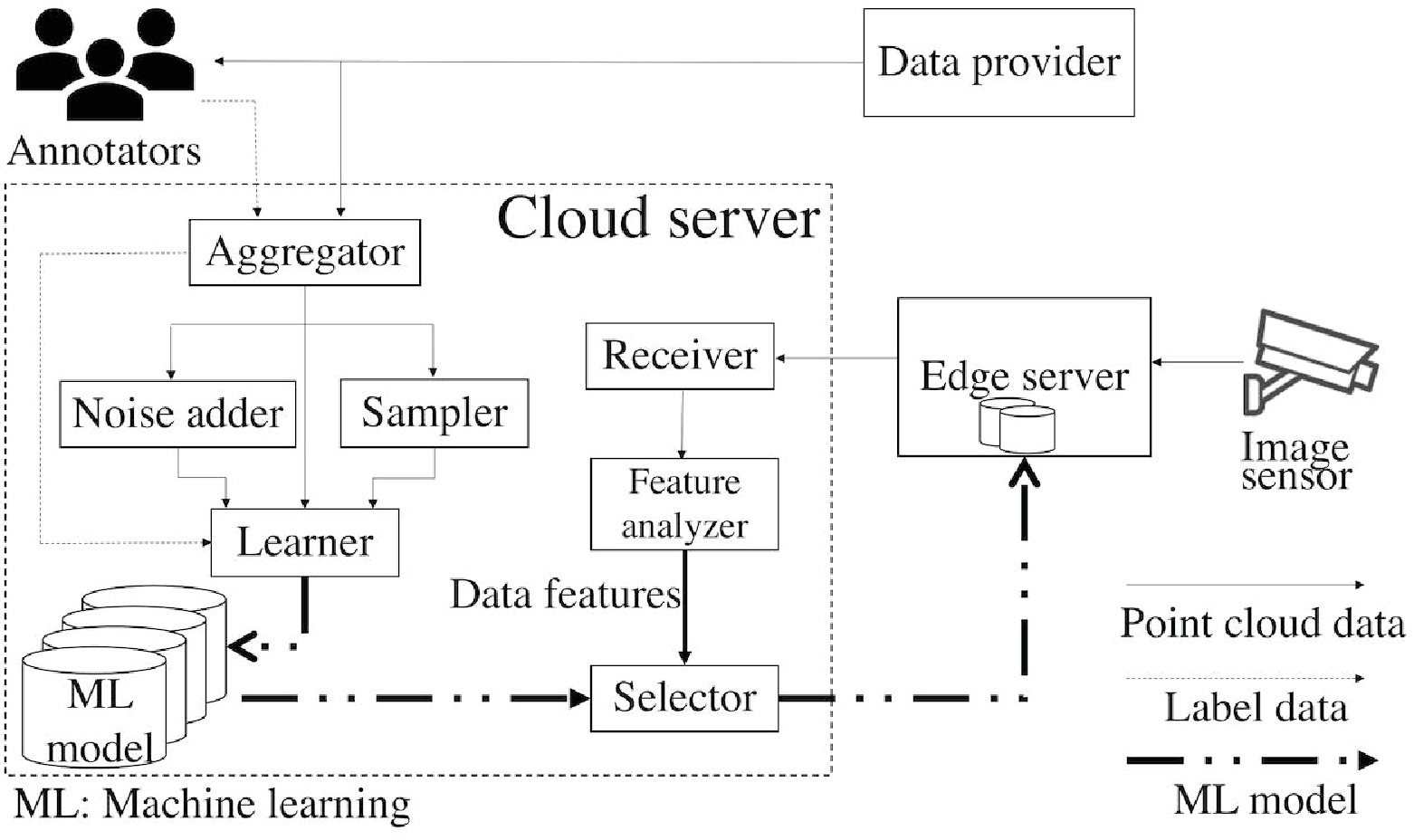}}
	\caption{Block diagram of system model. Thin solid arrows represent the passing of point cloud data, thin dotted arrows represent the passing of 3D label data, and thick dotted arrows represent the passing of a DL model.}
	\label{fig:System model}
\end{figure}

Fig.~\ref{fig:System model} shows the block diagram of the system model assumed in the proposed framework. The main components of this system are a cloud server, a 3D image sensor, and an edge server. The cloud server consists of an aggregator to collect training data, a sampler to thin out point clouds, a noise adder to add artificial noise to point clouds, a learner to train DL models, a receiver to receive point cloud data from the edge server, a feature analyzer to analyze the features of the point cloud, and a selector to choose a suitable DL model.

This system assumes that the 3D image sensor is installed at a fixed position near specific intersections or roads to acquire point cloud data. The 3D image sensor sends the acquired point cloud data to the edge server in the form of streaming data. In the inference stage, the point cloud data are input to a DL model placed on the edge server for object detection.

On the cloud server, multiple DL models are trained and the proposed framework selects a suitable DL model to be placed on the edge server. The point cloud data used for training the DL models are not the data acquired by the 3D image sensor in this system but rather high-quality data acquired from anywhere in the world. The data provider represents the source of such training data, and the aggregator of the cloud server receives the point cloud data from the data provider in the training stage. At the same time, the aggregator receives the label data necessary for training the DL models from the annotators, who manually create accurate 3D label data using any data useful for labeling, such as 2D image data with color in addition to point cloud data. Then, the aggregator links the point cloud data and label data to each other and stores them. Next, the aggregator replicates the training data and sends them to the sampler and the noise adder, in order to artificially produce data with low density of points and with noise, respectively. Then, the learner receives multiple sets of training data and combines them with multiple DL methods to train multiple DL models.

In the beginning or middle of the inference stage, the edge server sends the point cloud data acquired by the 3D image sensor and target data that indicate which class of objects is to be detected to the cloud server first. On the cloud server, the receiver receives the point cloud data and passes them to the feature analyzer. The feature analyzer analyzes the features of the point clouds of the received data (inference data), such as the spatial distribution of points, compares them with those of the original training data, and sends the comparison result (data features) to the selector. Then, the selector picks a suitable DL model based on the target data and the data features according to the predetermined procedure and sends it back to the edge server.

We summarize the operations of the training and inference stages in Algorithms~\ref{alg:train} and \ref{alg:inf}, respectively.

\begin{algorithm}[t]
	%		\small
	\caption{Training stage}
	\label{alg:train}
	\begin{algorithmic}[1]
		\renewcommand{\algorithmicrequire}{\textbf{Input:}}
		\renewcommand{\algorithmicensure}{\textbf{Output:}}
		\REQUIRE The point cloud data for training
		\ENSURE Multiple DL models
		\STATE The aggregator receives the point cloud data for training from the data provider, and the label data from the annotators. 
		\STATE The aggregator links the point cloud data and label data to each other and then stores them.
		\STATE The aggregator replicates the training data and sends them to the sampler and the noise adder. 
		\STATE The sampler and the noise adder artificially produce data with low density of points and with noise, respectively.
		\STATE The learner receives multiple sets of training data and combines them with multiple DL methods to train multiple DL models.
	\end{algorithmic}
\end{algorithm}

\begin{algorithm}[t]
	%		\small
	\caption{Inference stage}
	\label{alg:inf}
	\begin{algorithmic}[1]
		\renewcommand{\algorithmicrequire}{\textbf{Input:}}
		\renewcommand{\algorithmicensure}{\textbf{Output:}}
		\REQUIRE The point cloud data for inference
		\ENSURE Detection results (bounding boxes)
		\IF {a DL model placed on the edge server is determined}
		\STATE The edge server sends the point cloud data and target data to the cloud server.
		\STATE The receiver receives the point cloud data and passes them to the feature analyzer.
		\STATE The feature analyzer analyzes the features of the point cloud data and compares them with those of the original training data. 
		\STATE The feature analyzer sends the data features to the selector.
		\STATE The selector picks a suitable DL model based on the target data and the data features according to the predetermined procedure.
		\STATE The selector sends back the picked DL model to the edge server.
		\ENDIF
		\STATE The DL model placed on the edge server inputs point cloud data acquired by the 3D image sensor and performs object detection.
	\end{algorithmic}
\end{algorithm}

\subsection{Factors for multi-model selection } 
\label{sec:Multi-model}
This section explains the details of the feature-based DL model selection, which is the key component of the proposed framework. The features of the point cloud data acquired in the real world are diversified due to various factors. We focus here on three factors: object class, density of points, and noise. In order to deal with these factors, the proposed framework uses multiple models of methods with different features and selects them appropriately, rather than using only a single model. In the following part, for each factor, we discuss which model should be selected in accordance with the features.

\subsubsection{Object class} 
\label{sec: Object class}
In an actual intersection of roads, various types of moving objects can be expected to appear; the targets of smart monitoring systems include not only cars on the road but also pedestrians and cyclists. The biggest difference between these objects is their size. The size of pedestrians and cyclists is naturally smaller than that of cars, and even within the category of cars, the size of large vehicles such as buses and trucks is quite different from that of private cars. The smaller the object size is, the physically smaller the number of points belonging to the same object is, and the fewer features there are to represent each object. Therefore, small-size objects are generally more difficult to detect. We assume that each object has a different algorithm that would be most effective for detection. The proposed framework is designed to select the most suitable model for each object.

The difference described above stems from the difference in the distance between the object center and the point clouds on the object surface. Due to the nature of point clouds acquired by 3D image sensors, they are always distributed on the surface of each object. In the anchor-free strategy, which generates the boxes (i.e., proposals) directly from the foreground points, the farther the distance between each foreground point on the object surface and the center of the ground-truth box is, the larger the center regression error is. Shi \textit{et al.} \cite{shi2020points} presented a regression method to reduce this effect, but it is still not suitable for detecting large-size objects. For small-size objects, the center regression error is small and the anchor-free strategy works fine. On the other hand, the anchor-based strategy typically has a high detection accuracy and a small distance between the center of the anchor and the ground-truth box, since the boxes are generated by fitting the anchor all over the entire space. Therefore, the anchor-based strategy is suitable for detecting large-size objects. However, for small-size objects, there is a possibility that the objects may be overlooked, since the box is not generated directly from the foreground point in the anchor-based strategy. Thus, it is reasonable to select the anchor-based strategy when targeting large-size objects and the anchor-free strategy when targeting small-size objects. On the basis of the above, the proposed framework first selects the box generation strategy in accordance with the size of the object to be detected.

\subsubsection{Density of points} 
\label{sec:Density of points}
We next focus on the density of points as a spatial distribution, which is one of the features of the point cloud data. The density of points depends largely on the performance and set position of the 3D image sensor used for measurement. For example, according to the product guide, the Velodyne HDL-64E LIDAR has 64 vertical channels and a horizontal resolution of 0.08, and can acquire up to 1,300,000 points per second \cite{Velodyne}. However, a 3D image sensor with a performance this good is still expensive to install and cannot be used easily. In contrast, according to the product guide, the relatively inexpensive Velodyne VLP-16 LIDAR has 16 vertical channels, a horizontal resolution of 0.1, and can acquire up to 300,000 points per second \cite{Velodyne}. In this case, the difference in the number of vertical channels is particularly big, and the density of points is proportionally dropped to a quarter. In addition, even when the same 3D image sensor is used, the density of points physically drops when the set position is at a high altitude. Thus, such low density of points is a factor of the incompleteness of data addressed in this paper.

Drop in density of points decreases the detection accuracy. A possible specific cause of this is the poorness of spatial features. Since spatial features are determined by the features of points in the vicinity, the fewer points there are, the poorer the spatial features needed to determine the class and location of objects are. This leads to difficulty with object detection, the extent of which depends on the processing unit of the point clouds adopted by each method. The voxel-based process is generally considered effective and efficient for feature learning and box generation \cite{shi2020points}. However, the voxel-based (and pillar-based) process might not work in a situation where there are few points and each point is highly important. This is because the quantization of points may result in the loss of valuable features. In contrast, the point-based process to handle each point directly can use all the valuable points as they are, and thus makes good use of these features. Therefore, we assume that models created by the method that adopts the point-based process are robust against the drop in density of points. On the basis of the above, the proposed framework selects the processing unit of point clouds in accordance with the spatial distribution of the inference data.

\subsubsection{Noise} 
\label{sec:Noise}
The point clouds acquired by a 3D image sensor can often contain noise. The point clouds generated by sensors using image-based 3D reconstruction techniques, which are often used in indoor measurements, have a lot of noise and outliers mainly due to matching ambiguities or image imperfections\cite{wolff2016point}. On the other hand, the time-of-flight (TOF) sensor, which can be used outdoors, generates point clouds by directly measuring the distance of objects, and thus does not generate noise due to the former factor. However, even with this type of sensor, which is assumed to be used in this system, noise may occur due to environmental issues such as illumination, material reflectance, and so on \cite{javaheri2017subjective}. TOF sensors can basically be divided into two types: indirect TOF (ITOF) sensors and direct TOF (DTOF) sensors. LIDARs are categorized into the DTOF sensors that can measure long distances. In addition, there are different types of LIDAR, such as a scanning LIDAR and a flash LIDAR, with different measurement methods and different tolerance to noise caused by environmental issues\cite{padmanabhan2019modeling}. As described above, the severity of the noise varies depending on the mechanism of the sensor and environmental issues. When the noise is severe, the accuracy of object detection also generally decreases. Thus, such noise added to points is the other factor of the incompleteness of data addressed in this paper.

A possible specific cause of noise is the change in the relationship between each point. Noise shifts the coordinates of each point from original places, which changes the distribution of point clouds. This leads to inaccuracy of spatial features and difficulty of object detection, while the effect of this also depends on the processing unit of point clouds adopted by each method. The point-based process, which processes each point one by one, picks up all the changes in the relationship between points, so the effect of inaccuracies in spatial features is accentuated. On the other hand, voxel-based and pillar-based processes with quantization can mitigate these effects, since they interrupt the process of taking the average or maximum value of the features of included points during feature aggregation. Therefore, in contrast to the case of the low density of points, the models that adopt the voxel-based or pillar-based process are more robust against the noise than those that adopt the point-based process, which is taken into account in the proposed framework.

\subsubsection{Producing data with incompleteness for DL model creation} 
\label{sec:Adding incompleteness}
Machine learning, especially supervised learning, is built on the premise that the training data are sufficient representations of the inference data \cite{kurakin2016adversarial}. Therefore, the quickest way to improve the performance of a DL model is to prepare training data according to the spatial distribution of the inference data. The best way is of course to use the point cloud data acquired by the 3D image sensor in Fig.~\ref{fig:System model} as the training data. However, the labeling process, which is essential for training, is costly and time-consuming; generally, manual labeling has become a serious bottleneck as collecting and storing data have become easier and large amounts of training data are required \cite{nguyenincomplete}. In comparison, it is much easier to label only one good quality dataset and then directly manipulate the dataset to reproduce the incompleteness. Therefore, the proposed framework uses the training data for reproducing the incompleteness, such as low density of points and noise. In this way, the proposed framework can avoid collecting and labeling the training data on its own before installing the system, which significantly reduces the time spent on preparing the training data. The training time itself depends on the training time of the DL method selected by the proposed framework.

Mukhaimar \textit{et al.} suggested that training on data with certain inaccuracies can produce models that are resistant to those inaccuracies \cite{mukhaimar2019comparative}. On the other hand, they also noted that data inaccuracy is generally unpredictable and it is not advisable to use models that focus on specific inaccuracy. However, it is not a problem to prepare a model specialized for each incompleteness in the proposed framework, for two reasons: 1) as mentioned in Section~\ref{sec:Density of points}, the density of points is easy to predict since it is caused by the performance of the 3D image sensor and the installation position, and 2) in this system, the model placed in the edge server is not fixed, but is selected based on the analysis of which factor of incompleteness exists.

\subsection{DL model selection procedure}
\label{sec:Procedure}
\begin{figure*}[t]
	\centerline{\includegraphics[width=0.75\linewidth]{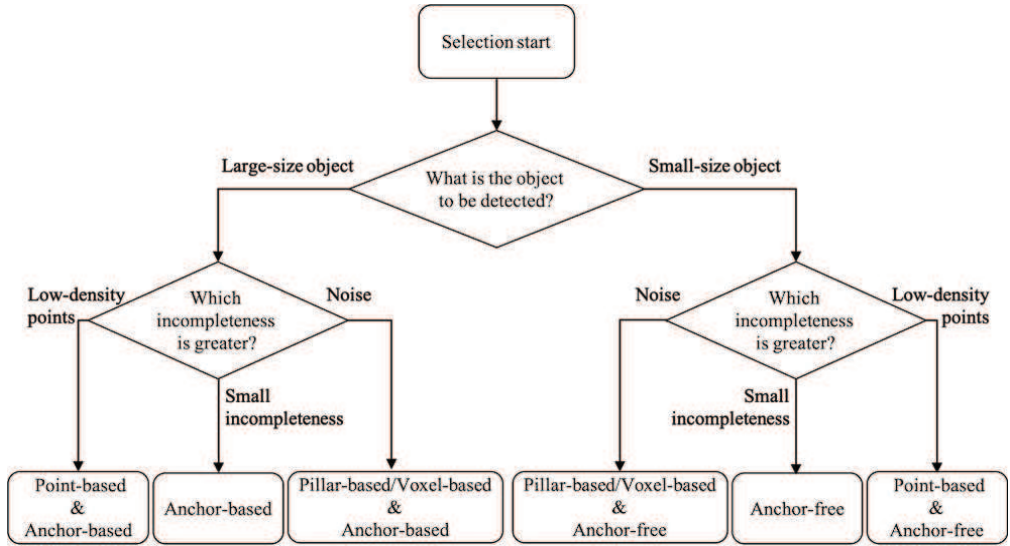}}
	\caption{DL-method selection flowchart.}
	\label{fig:Model selection}
\end{figure*}

This section explains the procedure for selecting the model of a suitable DL method to be sent to the edge server in the proposed framework, using the flowchart in Fig.~\ref{fig:Model selection}. As described in Section~\ref{sec:System model}, the selector selects a suitable DL method based on the features of the inference data analyzed by the feature analyzer. In general, DL models perform object detection in the following steps: extracting areas where objects may exist, classifying object categories according to predefined confidence thresholds, and refining their locations. As a result, information about the category and location of each object is output. With this, before selecting a DL model, it is necessary to decide the target object manually, i.e., which object is to be detected, though it is possible to change it during system operation. Then, the proposed framework uses a selected DL model to identify the location of the determined target object.

In Fig.~\ref{fig:Model selection}, for DL-method selection, the first branch regarding the box generation strategy occurs depending on the size of the target object. Specifically, it is narrowed down to DL methods that adopt the anchor-based strategy for large-size objects and the anchor-free strategy for small-size objects. Once that is done, the next branch regarding the processing unit of point clouds occurs depending on the type and degree of incompleteness of the inference data. Specifically, it is narrowed down to DL methods that adopt the point-based process when the density of points is low, and the pillar-based or the voxel-based process when noise is severe. However, it should be noted that, if the degree of incompleteness is small or there is a trained model using data with incompleteness close to the inference data, first, a DL method is selected only in accordance with the size of the target object. Then, at the next branch, the option after the arrow of `small incompleteness' is selected. After selecting the DL method, DL model selection is performed among the DL models trained by the selected DL method. At this time, the DL model using the training data which has the closest degree of incompleteness to the inference data is selected.

The detection time of the DL model depends on the DL method used for training. Therefore, the maximum detection time of the proposed framework is the slowest detection time among the candidate DL methods. If there is a requirement for the detection speed, the user needs to prepare multiple DL methods whose detection speed satisfies the requirement, and select one of them using a modified flowchart, which has an additional branch by detection speed in the first stage of Fig.~\ref{fig:Model selection}.

The flowchart in Fig.~\ref{fig:Model selection} helps us to determine which DL method should be used in accordance with the target object and the inference data. In reality, to implement DL-method selection based on this flowchart, we need to design quantitative selection criterion, i.e., which type of objects is categorized into large- or small-size object, what density of points is high or low, what level of noise is severe or not, and so on. The performance evaluation presented in Section~\ref{sec:Evaluation} suggests how to design such selection criterion. In addition, these evaluation results reveal the difference in detection accuracy for each feature and ensure the validity of the flowchart in Fig.~\ref{fig:Model selection}.

\subsection{Discussion about feature combination}
\label{sec:comb}
As the proposed framework tries to do, it is common in the field of object detection to use an appropriate DL method for a specific purpose of application, e.g., car detection. To this end, so far we discussed the appropriate feature of DL method for each feature of the point cloud data. On the other hand, it may be possible to take an approach that combines these features to build one DL method with multiple purposes of application. In this regard, however, it should be noted that some features are difficult to combine in one method at the same time. For example, in terms of processing units of point clouds, point-based and pillar-based processes have opposite features in terms of quantization granularity. Future research is awaited to examine whether their strengths will be compatible or cancel each other out when they are combined to expand the range of applicable point cloud data. We believe that it is worthwhile to combine the features of DL methods based on the above discussion in order to build one appropriate DL method for a specific purpose of application (e.g., car detection in noisy point cloud data).

\section{Evaluations} 
\label{sec:Evaluation}
This section presents our evaluations that show a suitable model varies depending on the measurement situation, in order to demonstrate the effectiveness of the proposed framework.

\subsection{DL methods} 
\label{sec:ML methods}
In this evaluation, we compared the detection performance of PointRCNN \cite{shi2019pointrcnn}, Part-A$\mathbf{^2} $ net \cite{shi2020points}, PV-RCNN \cite{shi2020pv}, SECOND \cite{yan2018second}, and PointPillars \cite{lang2019pointpillars}. For Part-A$\mathbf{^2} $ net, two frameworks are provided that adopt anchor-based and anchor-free box generation strategies, respectively; we treat them as two distinct methods denoted as Part-A$\mathbf{^2} $-anchor and Part-A$\mathbf{^2} $-free, respectively, following Shi \textit{et al.}'s work \cite{shi2020points}. The features of the above six methods are summarized in Table \ref{tab:ML features}. We adopt the six methods to exhaustively cover the features of DL methods to demonstrate the effectiveness of the proposed feature-based model selection framework. We consider that the evaluation using these six methods can capture the impact of DL method features on detection accuracy since the architectures adopted in these six methods are also representative ones commonly adopted in many of the later DL methods (e.g., voxel-based Region Proposal Network in SECOND \cite{yan2018second}). Note that these features are broad categories; having the same features does not necessarily mean that the algorithm is exactly the same. For example, although PointRCNN and PV-RCNN both perform a point-based process in the second stage, PointRCNN applies PointNet++ to raw point clouds in the same proposal simply to aggregate point-wise features, while PV-RCNN uses PointNet to aggregate the features of surroundings key points into grid points in each proposal. Among these six methods, Part-A$\mathbf{^2} $-free has the longest detection time, which is average 0.35 seconds per frame, as shown in \cite{otsu}; the maximum detection time of the proposed framework is about 0.35 seconds per frame in this case.

\begin{table}[t]
	\caption{Features of DL methods compared in evaluation. A dash in the second stage column indicates a one-stage method with no second stage.}
	\begin{center}
	\scriptsize
		\renewcommand{\arraystretch}{1.4}
%		\scalebox{0.92}{
			\begin{tabular}{|c|c|c|c|}
				\hline
				\multirow{2}{*}{\textbf{DL method}}&\multicolumn{3}{c|}{\textbf{Features}} \\
				\cline{2-4} 
				& \textbf{First stage}& \textbf{Second stage}& \textbf{Box generation strategy} \\
				\hline
				\textbf{PointRCNN} & Point-based & Point-based & Anchor-free \\
				\hline
				\textbf{Part-A$ ^2 $-free} & Voxel-based & Voxel-based & Anchor-free \\
				\hline
				\textbf{Part-A$ ^2 $-anchor} & Voxel-based & Voxel-based & Anchor-based \\
				\hline
				\textbf{PV-RCNN} & Voxel-based & Point-based & Anchor-based \\
				\hline
				\textbf{SECOND} & Voxel-based & \textbf{--} & Anchor-based \\
				\hline
				\textbf{PointPillars} & Pillar-based & \textbf{--} & Anchor-based \\
				\hline
			\end{tabular}
%		}
		\label{tab:ML features}
	\end{center}
\end{table}

\subsection{Dataset} 
\label{sec:Dataset}

\begin{figure}[t]
	\begin{center}
		\subfloat[Car]{
			\includegraphics[clip,width=0.7\linewidth]{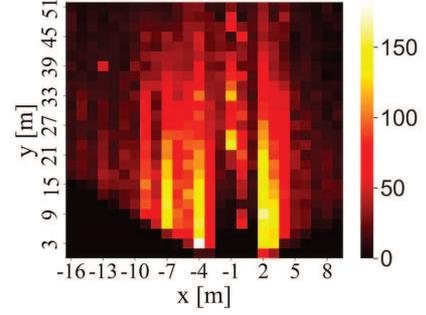}} \quad
		\subfloat[Pedestrian]{
			\includegraphics[clip, width=0.7\linewidth]{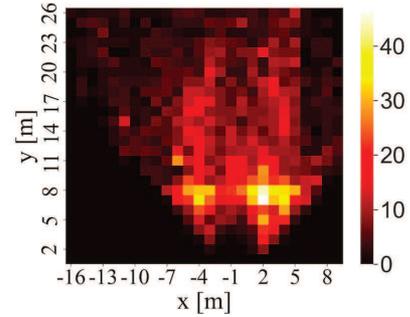}}
		\caption{Heat map showing distribution of objects in terms of positions.}
		\label{fig:heat}
	\end{center}	
\end{figure}

\begin{figure}[t]
	\begin{center}
		\subfloat[Car]{
			\includegraphics[clip,width=\linewidth]{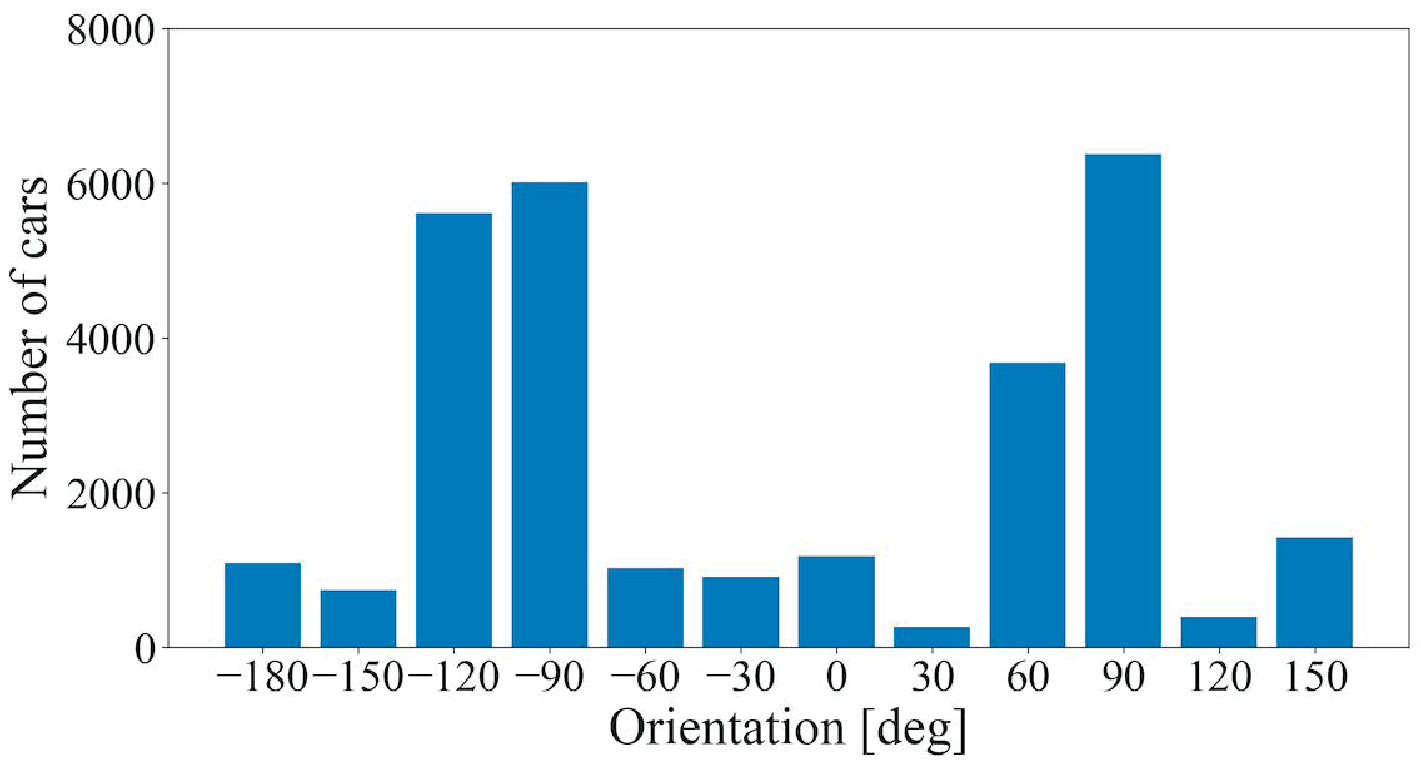}} 
		\\
		\subfloat[Pedestrian]{
			\includegraphics[clip, width=\linewidth]{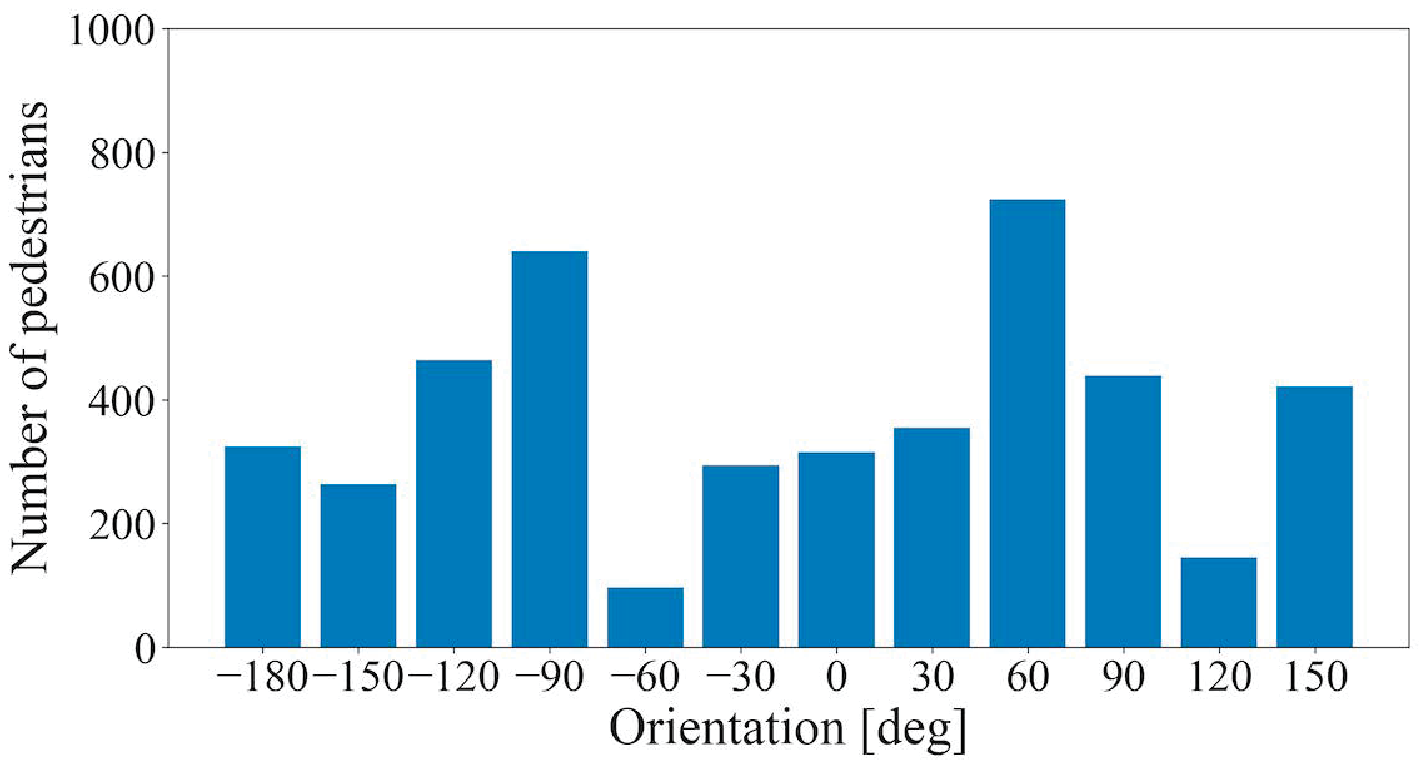}}
		\caption{Histogram showing distribution of objects in terms of orientations.}
		\label{fig:ori}
	\end{center}	
\end{figure}

\begin{figure}[t]
	\begin{center}
		\subfloat[Car]{
			\includegraphics[clip,width=\linewidth]{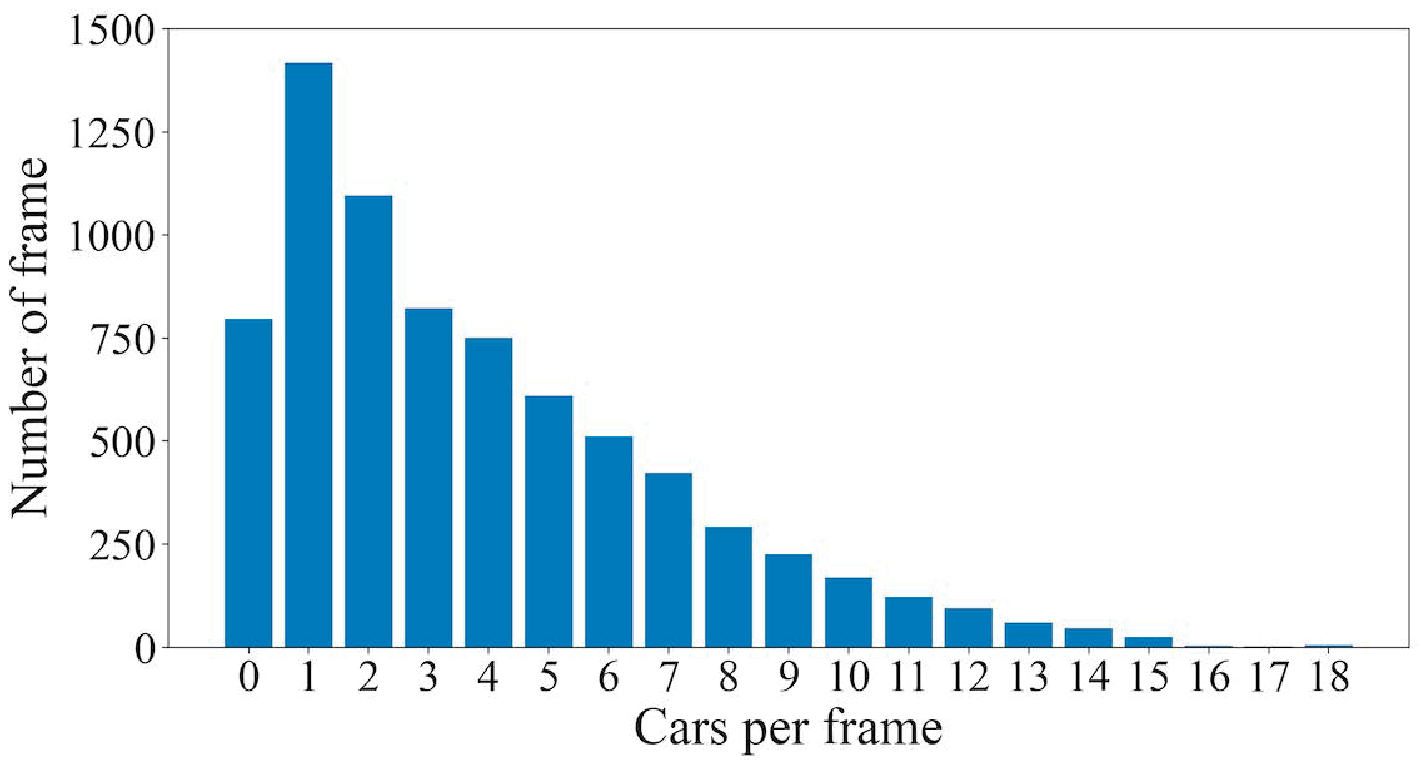}} 
		\\
		\subfloat[Pedestrian]{
			\includegraphics[clip, width=\linewidth]{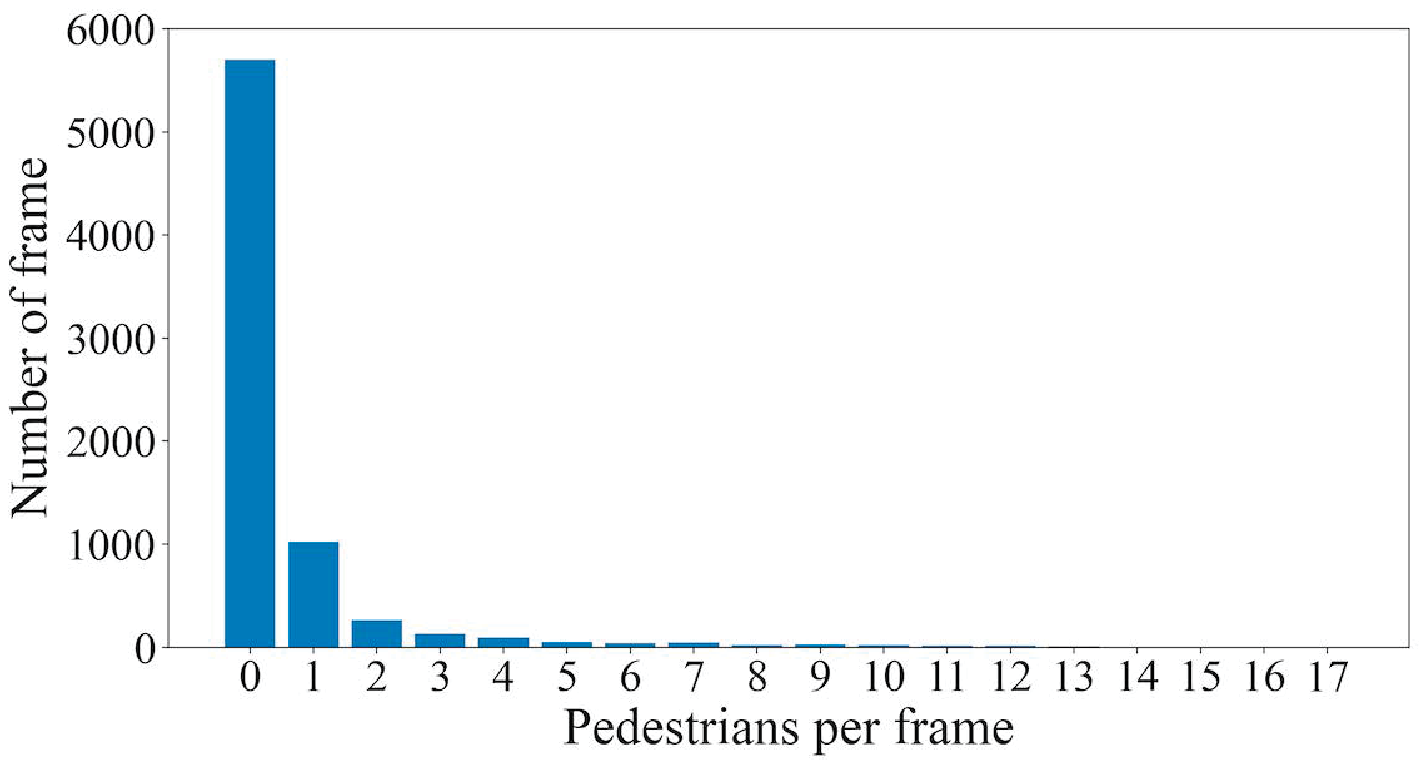}}
		\caption{Histogram showing distribution of objects in terms of number of objects per frame.}
		\label{fig:num}
	\end{center}
\end{figure}

We utilized the KITTI dataset \cite{geiger2012we} in this evaluation, which is often used to evaluate 3D object detection for autonomous driving. This dataset contains a training sample of 7481 frames and a test sample of 7518 frames. We used the training sample in this evaluation, as only the training sample provides the label data. The training sample is further divided into a training split of 3712 frames and a validation split of 3769 frames. We use the training split to train DL models and the validation split to evaluate the detection accuracy.

The point cloud data included in this dataset were acquired by a Velodyne HDL-64E LIDAR on the roof of a car while driving around a city in Germany. Therefore, this dataset is not time-series data that have been continuously acquired at a particular location. In addition, this dataset has a variety of occurrence patterns of objects. According to \cite{geiger2012we}, the frames in this dataset are selected so that it can be a diverse dataset with high entropy in the distribution of object orientation and number of non-occluded objects. Figs.~\ref{fig:heat}, \ref{fig:ori}, and \ref{fig:num} show the distributions of objects in terms of positions, orientations, and the numbers of objects, respectively, appearing in the training sample of 7481 frames. In Fig.~\ref{fig:heat}, we can see that objects are not only gathered on the lane or sidewalk, but also spread out to the front left and right. In Fig.~\ref{fig:ori}, we can also see that this dataset contains objects of various orientations, not limited to the orientation along the lane and sidewalk (i.e., $90^{\circ}$ and $-90^{\circ}$). Fig.~\ref{fig:num} shows that the distribution of the number of objects per frame has a wide tail. By using such a dataset to train the DL models, it is possible to generate a general model that can be applied to any road or intersection as it is, rather than a model specific to a certain location.

\subsection{Model implementation} 
\label{sec:Model implementation}
This section gives specific explanations about training the DL models. We prepared training data with pseudo incompleteness, as discussed in Section~\ref{sec:Adding incompleteness}. Here, we explain the sampling schemes to drop the density of points in Section~\ref{sec:Train sampling}, the way of adding noise to points in Section~\ref{sec:Train noise}, and the details of the training, including the various parameters we used, in Section~\ref{sec:Train detail}.

\subsubsection{Sampling of training data} 
\label{sec:Train sampling}
In this evaluation, two sampling schemes were used to reproduce the drop in density of points. First, a brief explanation of each scheme is given, followed by a discussion of the differences between the two and the reasons for their adoption.

\textbf{Voxel Grid Filter:} A commonly used scheme for sampling is the Voxel Grid Filter included in the point cloud library (PCL) \cite{rusu20113d}. This scheme first creates a 3D voxel grid by dividing the entire space into voxels of the specified size. After that, the average of each feature (x-y-z coordinates and intensity) of the points in the same voxel is calculated. Finally, all points in the voxel are deleted and a point with the previously calculated features is added in their place. As a result, there is one point at the centroid of each voxel.

\textbf{Uniform Sampling:} Another sampling scheme is Uniform Sampling, which is also included in the PCL library. This scheme is the same as the Voxel Grid Filter up to the stage of creating the 3D voxel grid. The difference is that, among the points included in the same voxel, only a point with the shortest distance to the center coordinate of the voxel is kept.

\textbf{Discussion about sampling:} Both Voxel Grid Filter and Uniform Sampling are capable of uniformly downsampling the entire point cloud. In general, Voxel Grid Filter is used for point cloud downsampling to lighten the processing. The advantage of this filter is that it does not affect the overall spatial distribution much, though it has the disadvantage of causing points to appear in locations where they should not. This is similar to the effect of noise, which means that a factor other than density of points may be involved. For this reason, our evaluation also uses Uniform Sampling, which leaves the originally existing point cloud for one point per voxel.

For simplicity, we used the number of points contained in each frame as the density of points of that frame. By setting the same voxel size for the two sampling schemes, the ratio of the number of points before sampling to after sampling (hereafter ``normalized number of points'') can be made equal in each frame. In this evaluation, three patterns with different lengths of one side of the voxel were prepared: 10 cm, 20 cm, and 40 cm. Due to the nature of the sampling schemes, while the normalized number of points per frame varied widely, the average values for all frames were about 0.478, 0.263, and 0.123, respectively. Hereafter, the models trained on the sampled data are called ``Voxel Grid 1/2 model,'' ``Voxel Grid 1/4 model,'' ``Uniform 1/8 model,'' etc. by approximating the normalized number of points as described above. We did not prepare a Voxel Grid 1/8 model for PV-RCNN or Part-A$\mathbf{^2} $-anchor, since they could not be trained due to program settings.

\subsubsection{Adding noise to training data} 
\label{sec:Train noise}
Noise in point clouds generated by 3D image sensors is typically modeled as additive white Gaussian noise with mean 0 \cite{javaheri2017subjective}\cite{gschwandtner2011blensor}. Therefore, the noise added to the training data is also Gaussian noise. In this evaluation, data with two patterns of Gaussian noise were prepared, where the mean is 0 and the standard deviation is 0.04 and 0.08. The models trained with each pattern are called ``Noise 0.04 model'' and ``Noise 0.08 model,'' respectively.

\subsubsection{Training details} 
\label{sec:Train detail}
A selection of the training parameters addressed here is shown in Table \ref{tab:Train parameter}. Basically, these parameters and all other parameters were determined according to the settings of OpenPCDet \cite{openpcdet2020}. Some slight changes to batch size were made to match the performance of the graphics processing unit (GPU) used (single NVIDIA GeForce RTX 2070 GPU with 8G memory). While seven models with different ways of sampling were prepared (refer to Section~\ref{sec:Train sampling}) for each method, the same values are set to all the parameters for each model.

\begin{table}[t]
	\caption{Training parameters of each method.}
	\begin{center}
		\renewcommand{\arraystretch}{1.4}
		\begin{tabular}{|c|c|c|c|}
			\hline
			\multirow{2}{*}{\textbf{DL method}}&\multicolumn{3}{c|}{\textbf{Training parameters}} \\
			\cline{2-4} 
			& \textbf{Batch size}& \textbf{Leaning rate}& \textbf{Epochs} \\
			\hline
			\textbf{PointRCNN} & 2 & 0.01 & 80 \\
			\hline
			\textbf{Part-A$ ^2 $-free} & 4 & 0.003 & 80 \\
			\hline
			\textbf{Part-A$ ^2 $-anchor} & 2 & 0.01 & 80 \\
			\hline
			\textbf{PV-RCNN} & 1 & 0.01 & 80 \\
			\hline
			\textbf{SECOND} & 4 & 0.003 & 80 \\
			\hline
			\textbf{PointPillars} & 4 & 0.003 & 80 \\
			\hline
		\end{tabular}
		\label{tab:Train parameter}
	\end{center}
\end{table}

\subsection{Evaluation method} 
\label{sec:Evaluation strategy}

The KITTI dataset contains multiple moving objects, including cars, vans, trucks, trams, pedestrians, and cyclists. Among them, we focus on detecting cars, pedestrians, and cyclists in this evaluation. This is because these objects are commonly used as targets for performance evaluations using this dataset, as used in \cite{lang2019pointpillars} \cite{shi2019pointrcnn}. Note that the proposed framework can be applied for other objects since it makes decisions based on the size of the object.

We use Average Precision (AP) with a rotated 3D Intersection over Union (IoU) threshold of 0.7 for cars and 0.5 for pedestrians and cyclists as the evaluation metric, which is also used in the KITTI benchmark \cite{kittibenchmarks}. Note that KITTI has several evaluation benchmarks, such as 3D, BEV, and AOS, among which this evaluation uses the 3D. While there are several ways to calculate AP \cite{padilla2020survey}, we chose the type that calculates it based on 40-point recall positions, as in \cite{simonelli2019disentangling}.

Sampling and adding noise to the validation data are also performed to simulate various measurement situations. Mukhaimar \textit{et al.} \cite{mukhaimar2019comparative} used Random Sampling to reduce the number of points to evaluate the robustness of DL methods for 3D shape recognition for low density of points. Similarly, this evaluation used Random Sampling to reproduce the data with low density of points with a normalized number of points of 0.5, 0.25, and 0.125. For noise, Gaussian noise with mean 0 was used as in Section~\ref{sec:Train noise}. This evaluation reproduced the data with Gaussian noise with standard deviations of 0.02, 0.04, 0.06, 0.08, and 0.1.

\subsection{Results} 
\label{sec:Results}
This section reports the evaluation results of each model for data simulating various measurement situations. In the following, the models trained on the original training data without sampling are called the ``Original model,'' and the original validation data are called the ``original data.'' Note that all accuracies in Figs. \ref{fig:Results density} to \ref{fig:Results Uniform} are results of the moderate class, which is the class of detection difficulty defined by the KITTI benchmark.

\subsubsection{Comparison of models created by different DL methods for each object class} 
\label{sec:Results object}

\begin{table*}[t]
	\caption{Performance comparison of Original models of each method using original data. All numbers in the table represent AP (\%), and bold numbers indicate the maximum value in each column. Easy, Moderate, and Hard represent the detection difficulty as defined by the KITTI benchmark \cite{kittibenchmarks}.}
	\begin{center}
		\renewcommand{\arraystretch}{1.4}
		\begin{tabular}{|c|c|c|c|c|c|c|c|c|c|}
			\hline
			\multicolumn{1}{|l|}{} & \multicolumn{3}{c|}{\textbf{Car}} & \multicolumn{3}{c|}{\textbf{Pedestrian}} & \multicolumn{3}{c|}{\textbf{Cyclist}} \\ \hline
			\textbf{DL method} & \multicolumn{1}{c|}{\textbf{Easy}} & \multicolumn{1}{c|}{\textbf{Moderate}} & \multicolumn{1}{c|}{\textbf{Hard}} & \multicolumn{1}{c|}{\textbf{Easy}} & \textbf{Moderate} & \multicolumn{1}{c|}{\textbf{Hard}} & \multicolumn{1}{c|}{\textbf{Easy}} & \multicolumn{1}{c|}{\textbf{Moderate}} & \multicolumn{1}{c|}{\textbf{Hard}} \\ \hline
			\textbf{PointRCNN} & 89.3 & 80.1 &78.0 & 61.8 & 54.6 & 47.9 & \textbf{92.4} & 72.8 & 68.9 \\ \hline
			\textbf{Part-A$ ^2 $-free} & 91.6 & 80.0 & 77.8 & \textbf{68.2} & \textbf{62.1} & \textbf{55.9} & 88.7 & \textbf{72.9} & \textbf{68.9} \\ \hline
			\textbf{Part-A$ ^2 $-anchor} & \textbf{92.3} & 82.2 & 79.8 & 61.6 & 53.0 & 47.2 & 90.2 & 72.7 & 68.7 \\ \hline
			\textbf{PV-RCNN} & 92.2 & \textbf{82.8} & \textbf{80.5} & 55.7 & 48.1 & 44.1 & 88.4 & 72.3 & 67.8 \\ \hline
			\textbf{SECOND} & 88.4 & 78.8 & 75.8 & 52.4 & 47.0 & 42.6 & 81.9 & 66.8 & 62.1 \\ \hline
			\textbf{PointPillars} & 86.7 & 75.7 & 72.6 & 50.4 & 43.7 & 39.0 & 82.1 & 62.9 & 59.0 \\ \hline
		\end{tabular}
		\label{tab:Resultus object}
	\end{center}
\end{table*}

Table \ref{tab:Resultus object} summarizes the detection accuracy of the Original model of each method using the original data. The methods in the third and fourth rows are the ones that adopt the anchor-free strategy, and the others are the ones that adopt the anchor-based strategy. When we look at the bold numbers representing the maximum value in each column, it is clear that the anchor-based strategy was superior for detecting cars, which are large-size objects, and the anchor-free strategy was superior for detecting pedestrians and cyclists, which are small-size objects. This result confirms that the DL method that achieves the highest detection accuracy changes depending on the target object size, and that the proposed framework, which selects the box generation strategy in accordance with the size of the object to be detected, works effectively. 

\subsubsection{Comparison of models created by different DL methods for density of points} 
\label{sec:Results density}

\begin{figure}[t]
	\centering
	\subfloat[Car]{\includegraphics[clip, width=\linewidth]{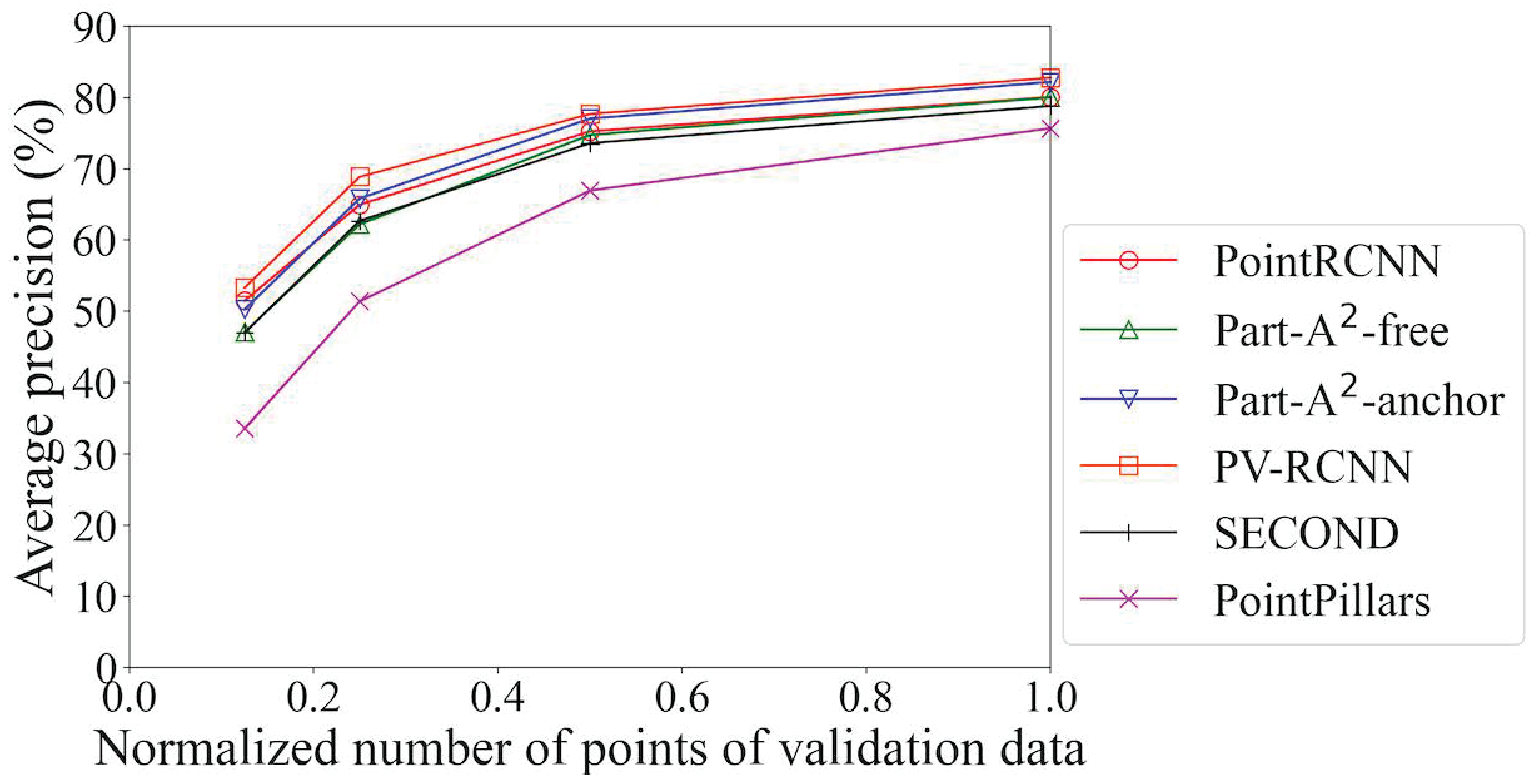}}
	\\
	\vspace{-12pt}
	\subfloat[Pedestrian]{\includegraphics[clip, width=\linewidth]{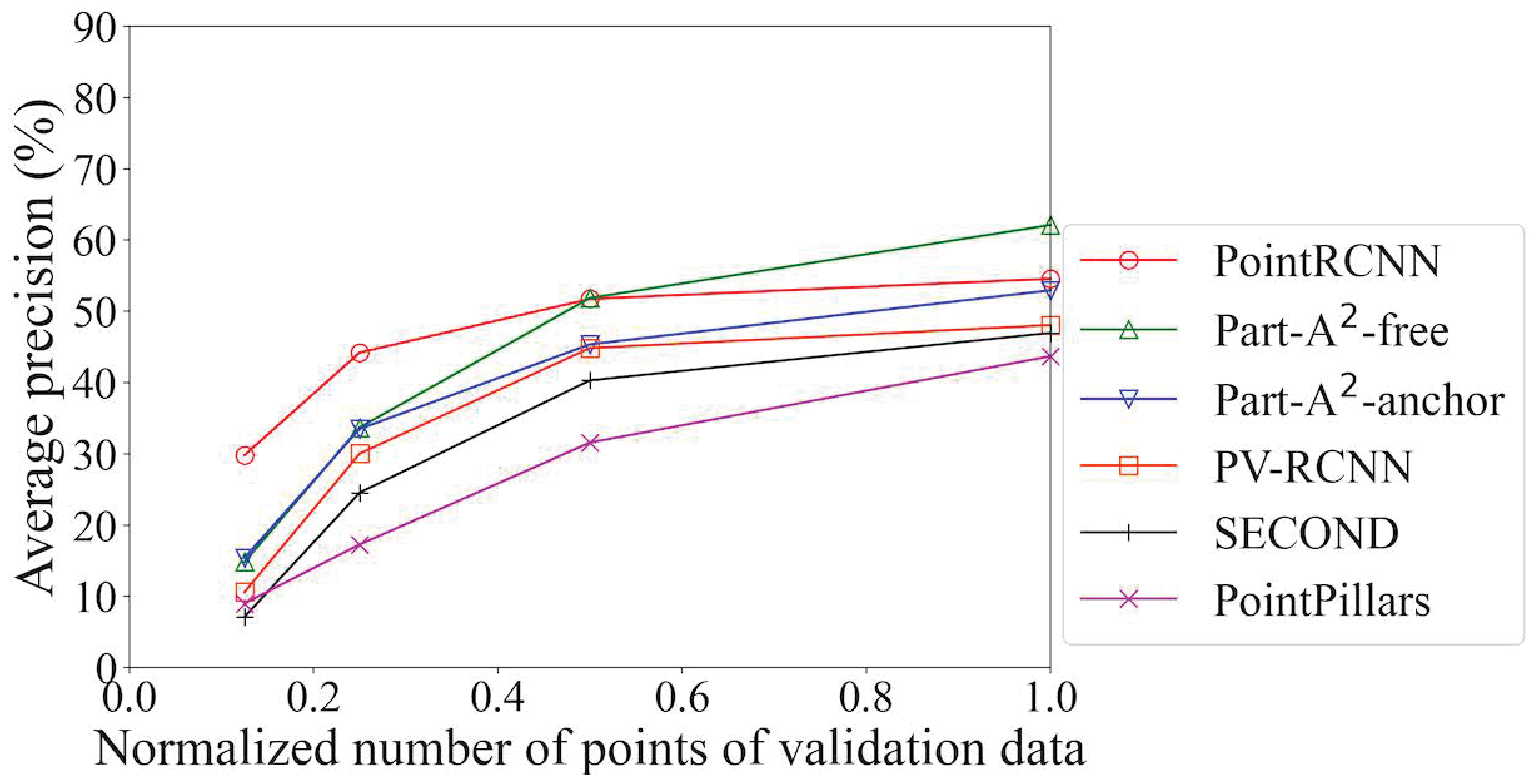}}
	\\
	\vspace{-12pt}
	\subfloat[Cyclist]{\includegraphics[clip, width=\linewidth]{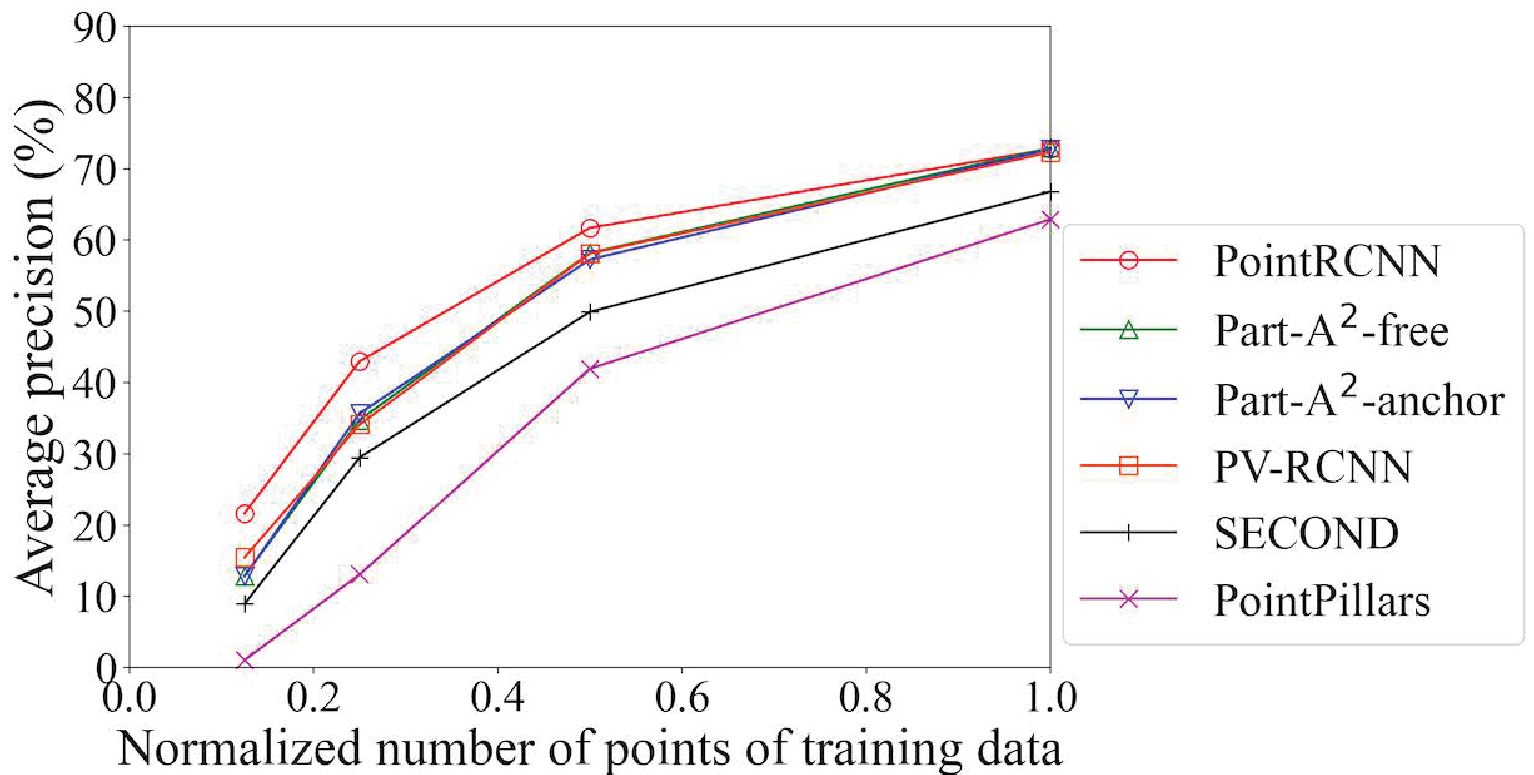}}
	
	\caption{Detection accuracy of each class of Original models for density of points. Horizontal axis represents the normalized number of points of the validation data, where 1.0 refers to the original data.}
	\label{fig:Results density}
\end{figure}

\begin{figure}[t]
	\centering
	\subfloat[Car]{\includegraphics[clip, width=\linewidth]{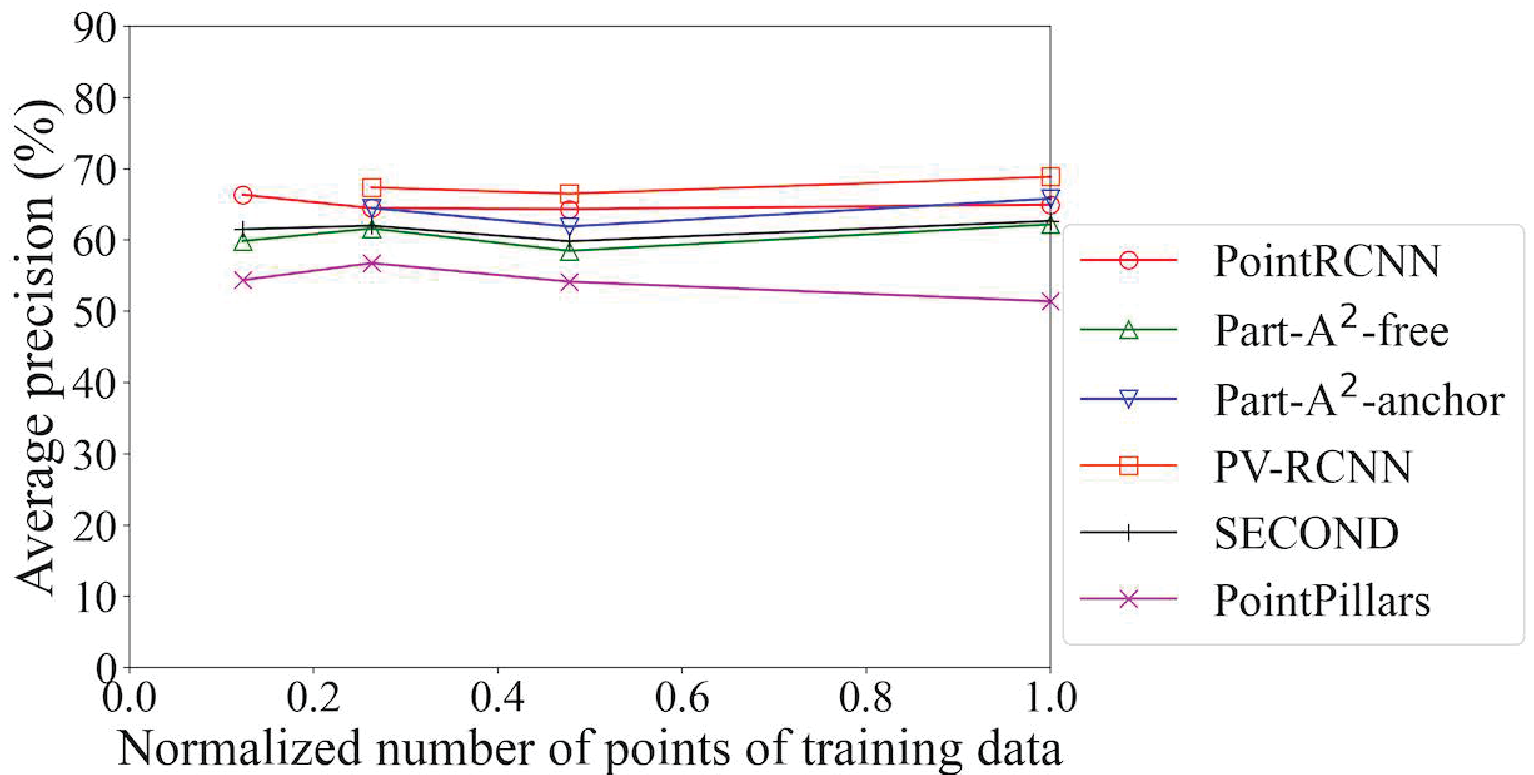}\label{subfig:Results Voxel Grid car}}
	\\
	\vspace{-12pt}
	\subfloat[Pedestrian]{\includegraphics[clip, width=\linewidth]{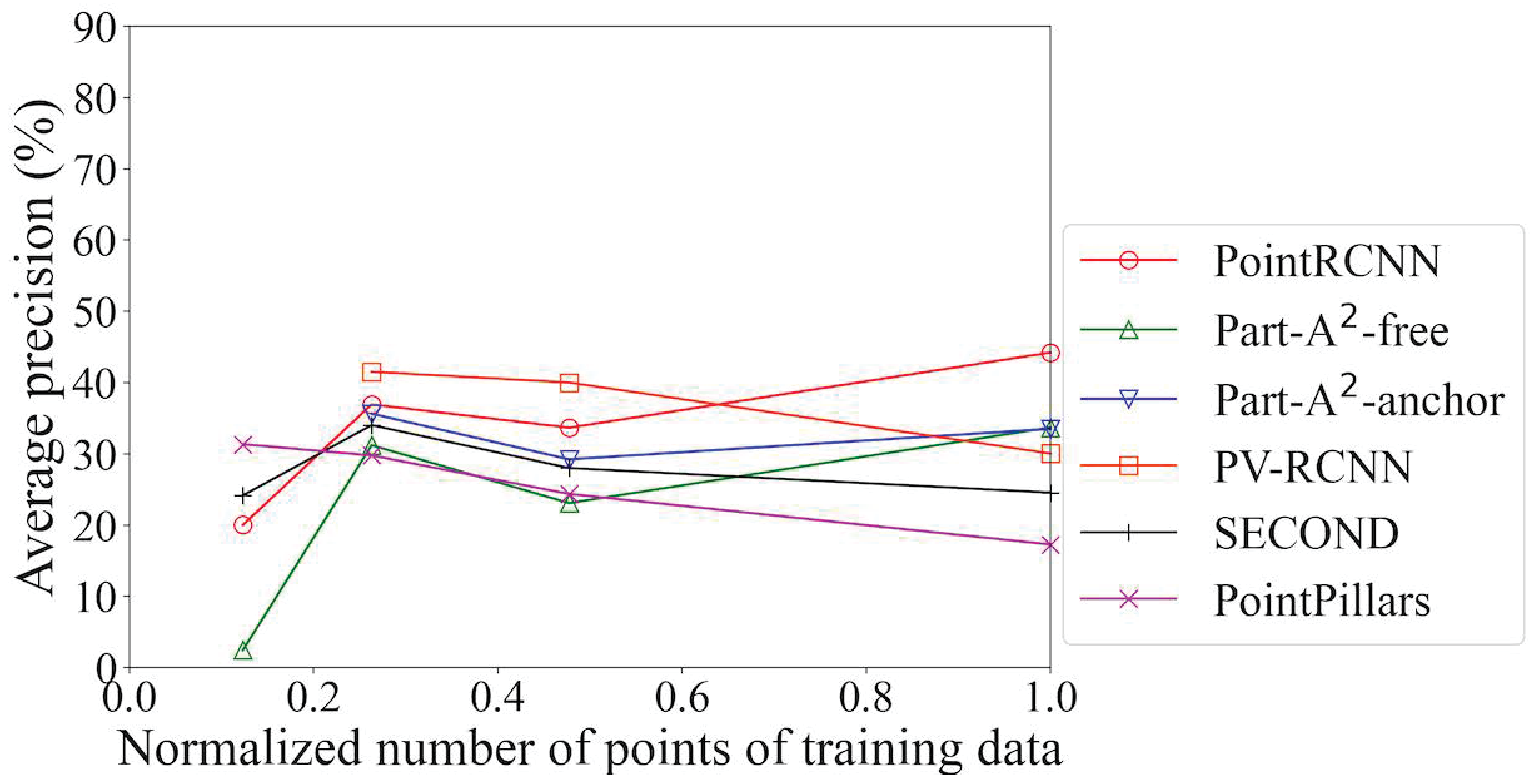}}
	\\
	\vspace{-12pt}
	\subfloat[Cyclist]{\includegraphics[clip, width=\linewidth]{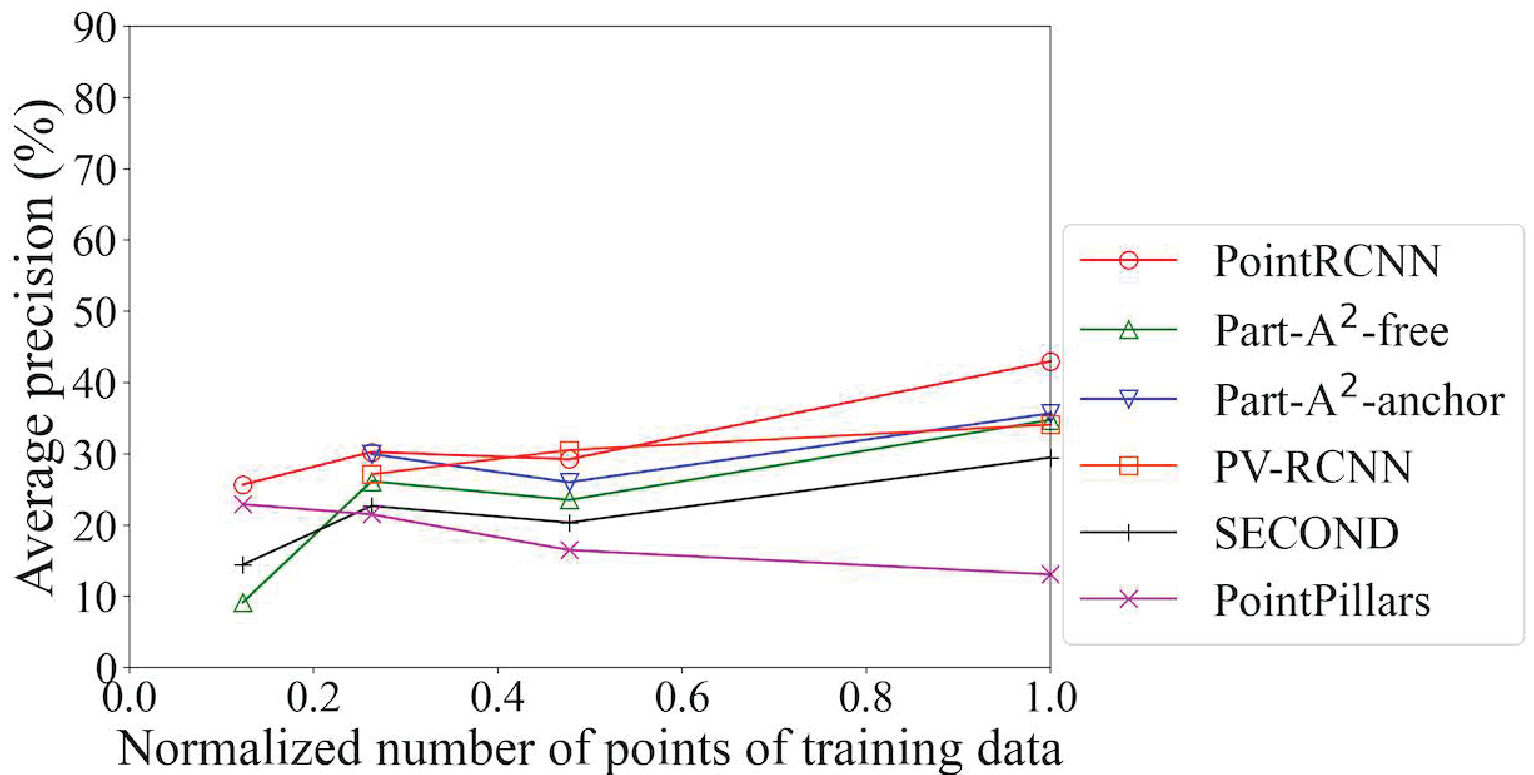}}
	
	\caption{Detection accuracy of Voxel Grid models for data with normalized number of points of 0.25. Horizontal axis represents the normalized number of points of the training data, where 1.0 refers to the Original model.}
	\label{fig:Results Voxel Grid}
\end{figure}

Fig.~\ref{fig:Results density} shows the detection accuracy of the Original models of each method for density of points. Two of the six methods, PointRCNN and PV-RCNN, adopt the point-based process. We can see here that the accuracy of the methods that adopt the point-based process was higher than the others in all classes for data with fewer points, especially for data with the normalized number of points of 0.25 and 0.125. For example, the accuracy of car detection for data with the normalized number of points of 0.125 can differ by up to 20\% (PointRCNN with point-based process vs. PointPillars with pillar-based process). In contrast, the accuracy of the methods that adopt the voxel-based or pillar-based process decreased sharply. Especially for pedestrians and cyclists, we can see that Part-A$ ^2 $-free, which had the maximum accuracy for the original data, was overtaken by the methods that adopt the point-based process as the drop in density of points became more severe. This result confirms that the DL method that achieves the highest detection accuracy changes depending on the density of points, and that the proposed framework, which selects the processing unit of point clouds in accordance with the spatial distribution of the inference data, works effectively. In addition, since PointRCNN adopts the anchor-free strategy and PV-RCNN adopts the anchor-based strategy, it is confirmed that the selection of box generation strategy by object size (described in Section~\ref{sec:Results object}) is compatible even for data with low density of points.

\subsubsection{Comparison of models created with different sampling rates} 
\label{sec:Results sampling model}

\begin{figure}[t]
	\centering
	\subfloat[Car]{\includegraphics[clip, width=\linewidth]{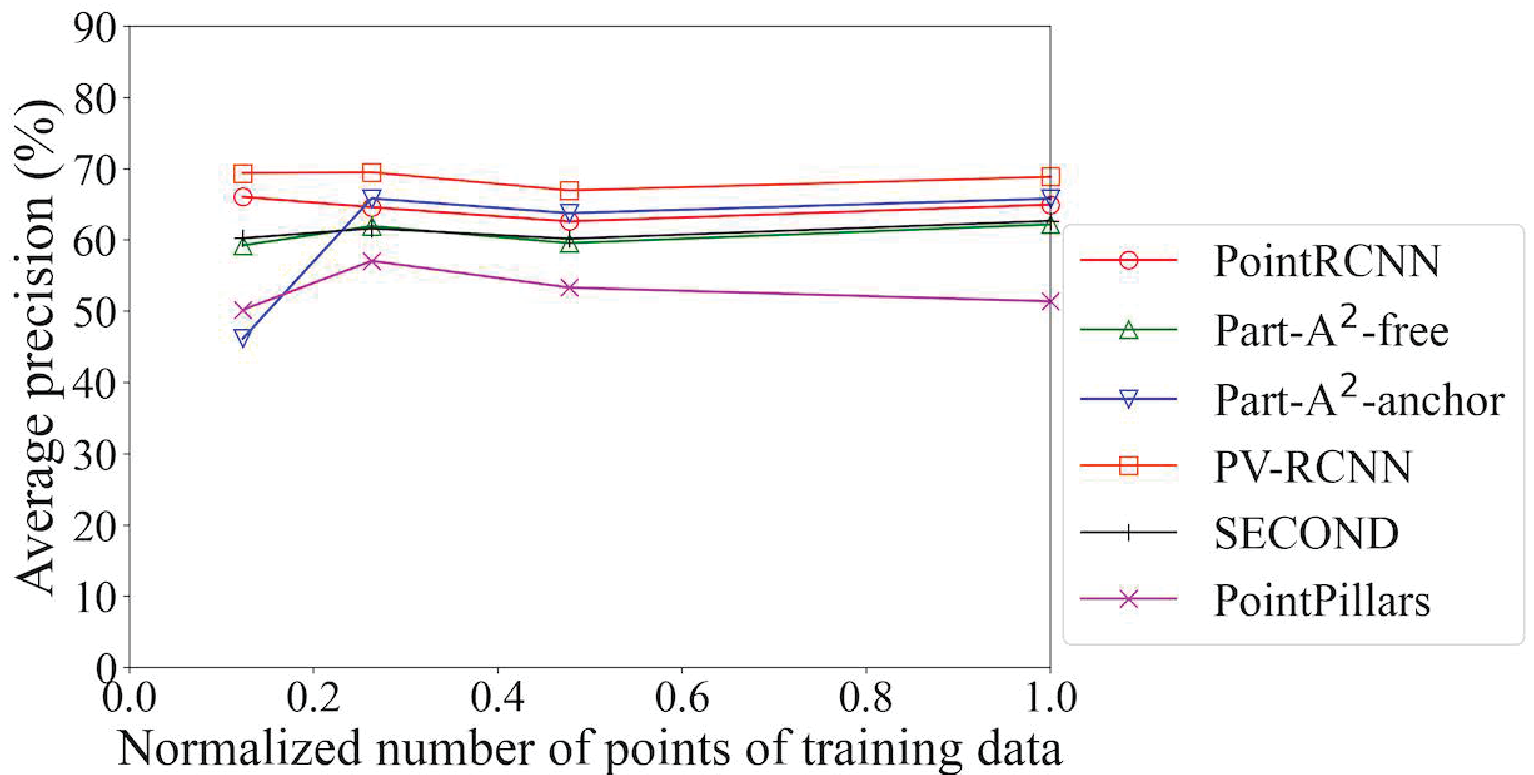}\label{subfig:Results Uniform car}}
	\\
	\vspace{-12pt}
	\subfloat[Pedestrian]{\includegraphics[clip, width=\linewidth]{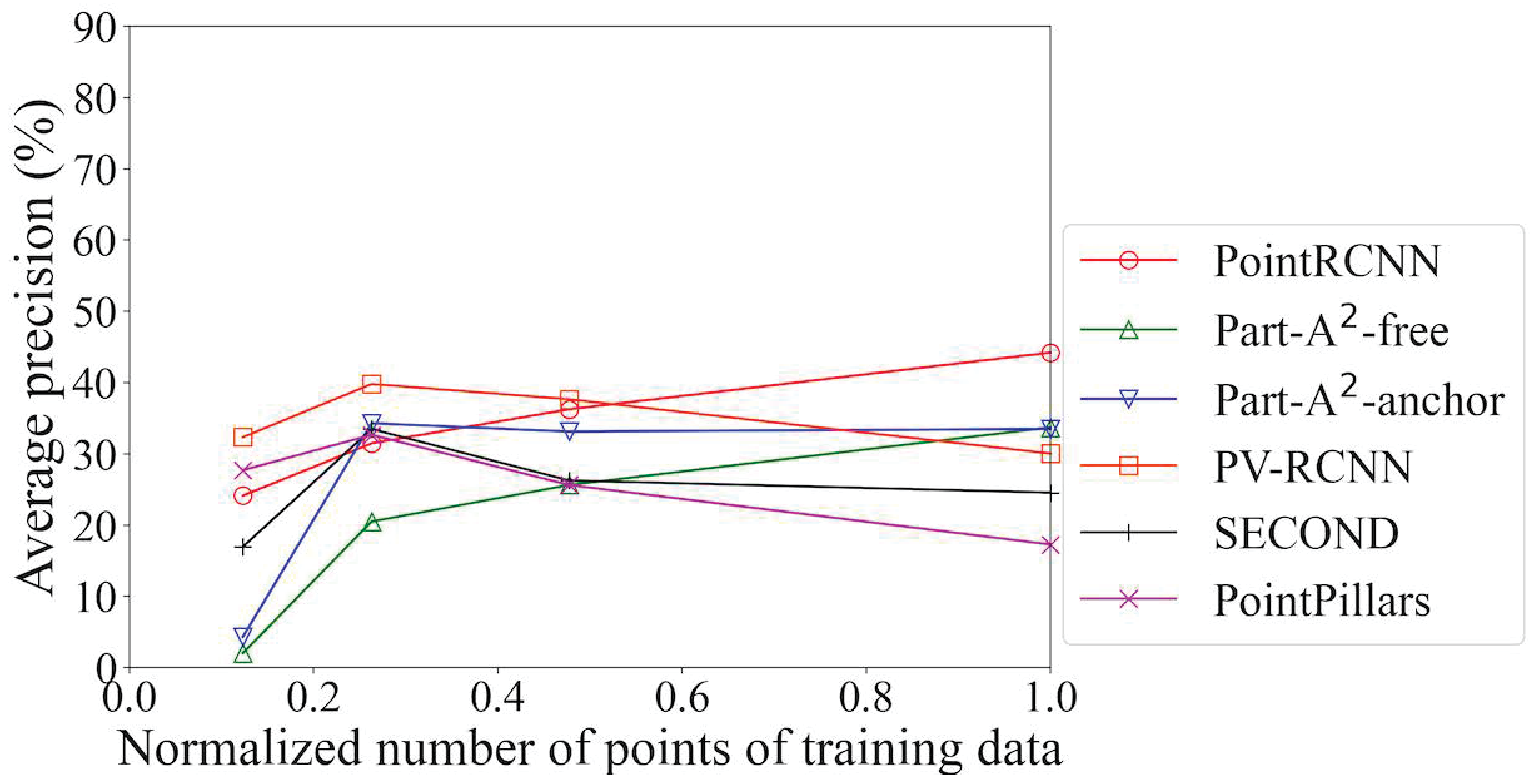}}
	\\
	\vspace{-12pt}
	\subfloat[Cyclist]{\includegraphics[clip, width=\linewidth]{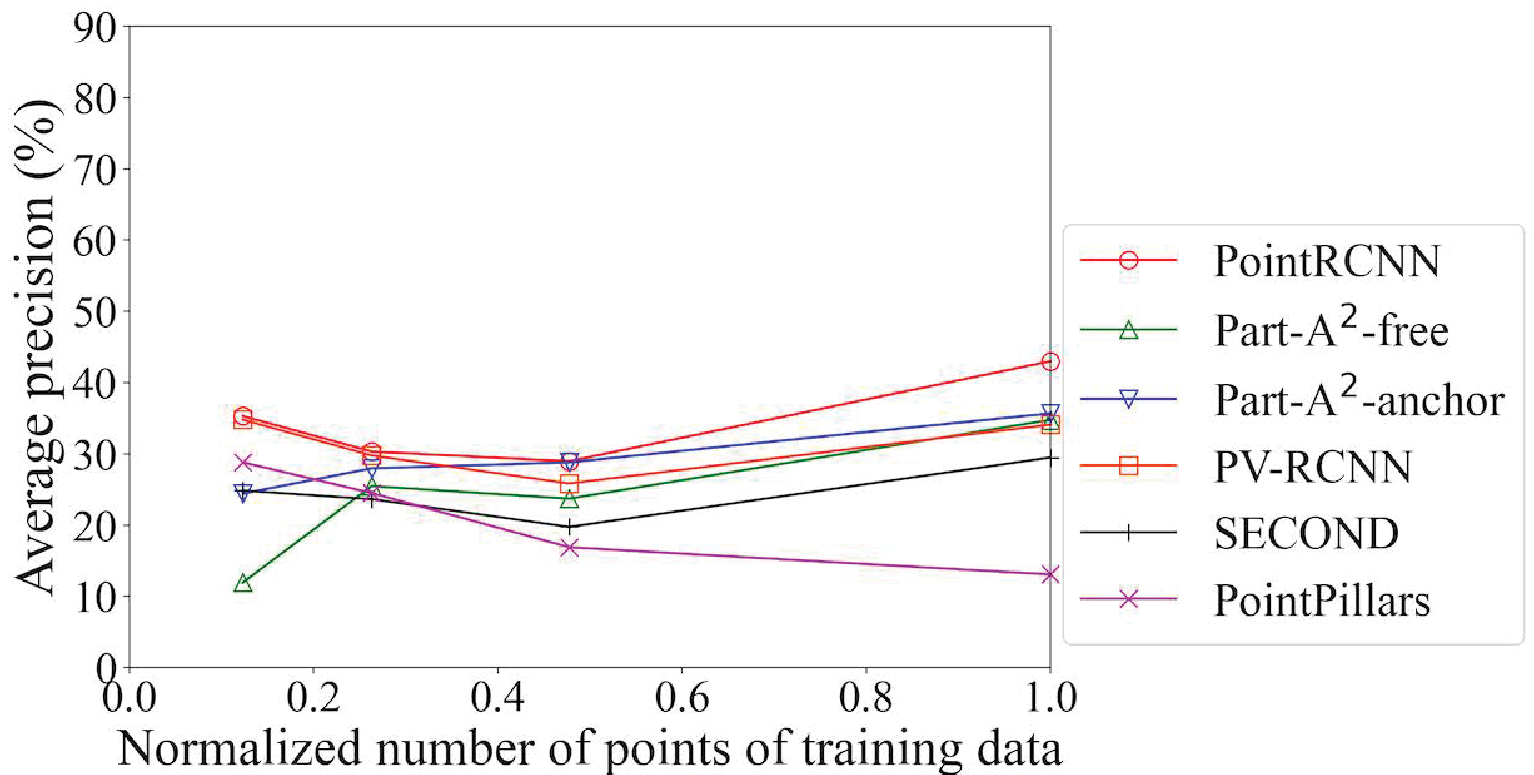}}
	
	\caption{Detection accuracy of Uniform models for data with normalized number of points of 0.25. Horizontal axis represents the normalized number of points of the training data, where 1.0 refers to the Original model.}
	\label{fig:Results Uniform}
\end{figure}

\begin{table}[bt]
	\caption{List of models that achieved maximum accuracy for data with normalized number of points of 0.25. For each method, the model with the maximum accuracy in each class is summarized out of seven models (including three Voxel Grid models, three Uniform models, and the Original model).}
	\begin{center}
		\renewcommand{\arraystretch}{1.4}
		\begin{tabular}{|c|c|c|c|}
			\hline
			\textbf{DL method} & \textbf{Car} & \textbf{Pedestrian} & \textbf{Cyclist} \\ \hline
			\textbf{PointRCNN} & Voxel Grid 1/8 & Original & Original \\ \hline
			\textbf{Part-A$ ^2 $-free} & Original & Original & Original \\ \hline
			\textbf{Part-A$ ^2 $-anchor} & Uniform 1/4 & Voxel Grid 1/4 & Original \\ \hline
			\textbf{PV-RCNN} & Uniform 1/4 & Voxel Grid 1/4 & Uniform 1/8 \\ \hline
			\textbf{SECOND} & Original & Voxel Grid 1/4 & Original \\ \hline
			\textbf{PointPillars} & Uniform 1/4 & Uniform 1/4 & Uniform 1/8 \\ \hline
		\end{tabular}
		\label{tab:Resultus sampling model}
	\end{center}
\end{table}

Figs.~\ref{fig:Results Voxel Grid} and \ref{fig:Results Uniform} show the detection accuracy of models created with sampling data (sampling models) for data with low density of points. Fig.~\ref{fig:Results Voxel Grid} is related to the Voxel Grid models and Fig.~\ref{fig:Results Uniform} is related to the Uniform models, both using the validation data with the normalized number of points of 0.25. Table \ref{tab:Resultus sampling model} summarizes the models with the maximum accuracy for each method and each class. As we expected, basically, the accuracy is improved by bringing the normalized number of points in the training data closer to that of the inference data (0.25 in this case). It is an exception that, for the methods in the second and third rows that adopt the anchor-free strategy, the Original model achieved the maximum accuracy in most cases. This is because this strategy generates the boxes directly from the foreground points; the benefit of making the spatial distribution of the training data similar to the inference data was small. 
When we look at the detection accuracy of cars of the models created by methods using the anchor-based strategy in Figs. \ref{subfig:Results Voxel Grid car} and \ref{subfig:Results Uniform car}, i.e., Part-A$\mathbf{^2} $ net, PV-RCNN, SECOND, and PointPillars, we see that the Uniform model achieved the maximum accuracy in three out of four methods, while the Voxel Grid model had lower accuracy than the Original model in three out of four methods. These results confirm that, basically, the selection procedure in the proposed framework can work effectively. They also suggest that the Uniform model is used for the large-size objects for which the anchor-based strategy is effective, while the Original model is used for the small-size objects for which the anchor-free strategy is effective.

Here, the effect of the proposed framework is reviewed in detail. Our proposed framework is intended to use the appropriate DL model depending on the situation. Therefore, we refer to the use of a single DL model for all situations as the benchmark for comparison. As an example, let the PointRCNN Original model, which has the highest average precision for pedestrians and cyclists in the original data in Table~\ref{tab:Resultus object}, be the DL model used in the benchmark. For example, in the detection of cars for data with low density of points shown in Fig.~\ref{subfig:Results Uniform car}, the proposed framework selects PV-RCNN according to the flowchart in Fig.~\ref{fig:Model selection}. In this case, the proposed framework improves the average precision by 4.5\% compared to the benchmark.

\subsection{Model comparison for data with noise} 
\label{sec:Results noise}

\begin{figure}[t]
	\centering
	\subfloat[Car]{\includegraphics[clip, width=\linewidth]{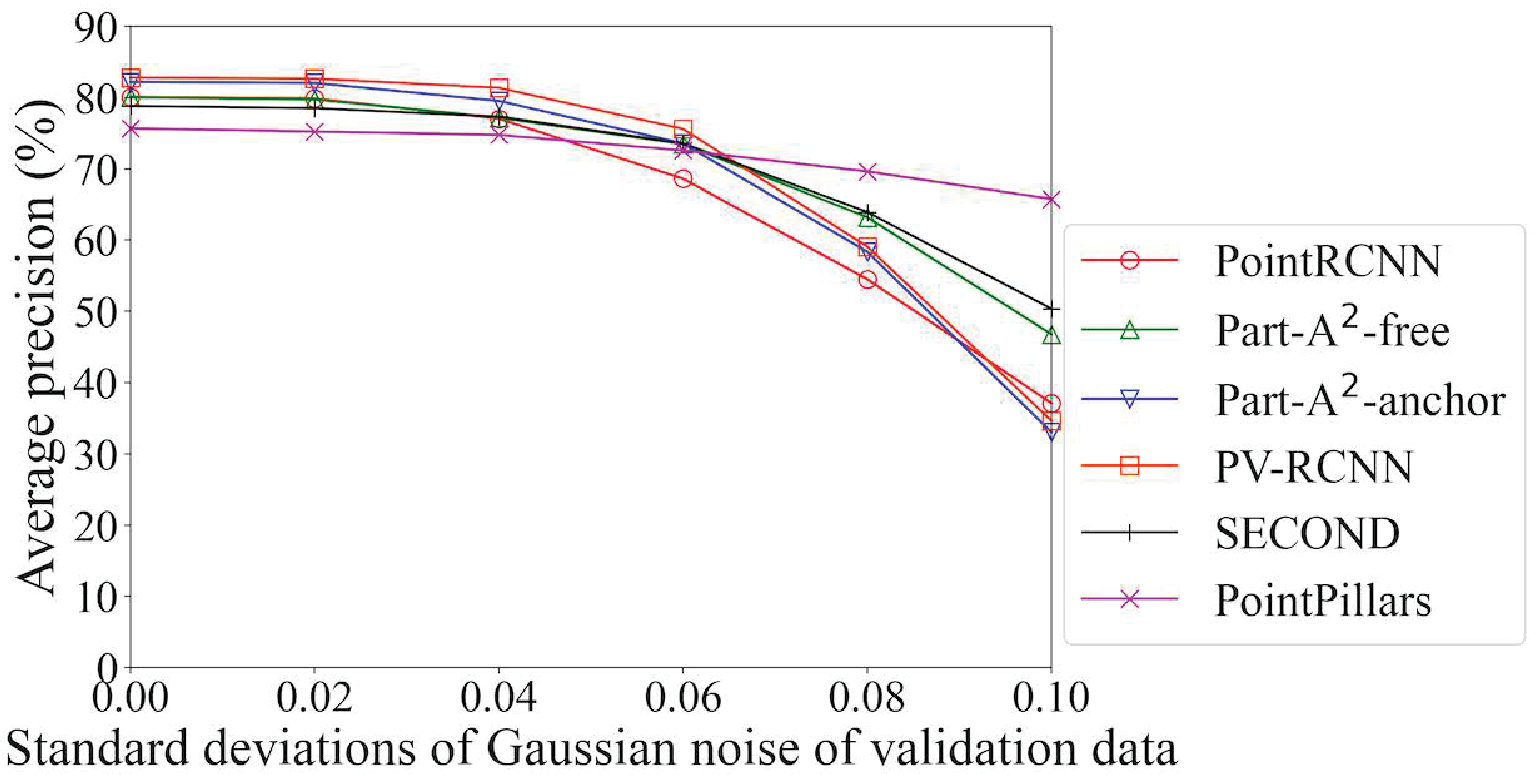}}
	\\
	\subfloat[Pedestrian]{\includegraphics[clip, width=\linewidth]{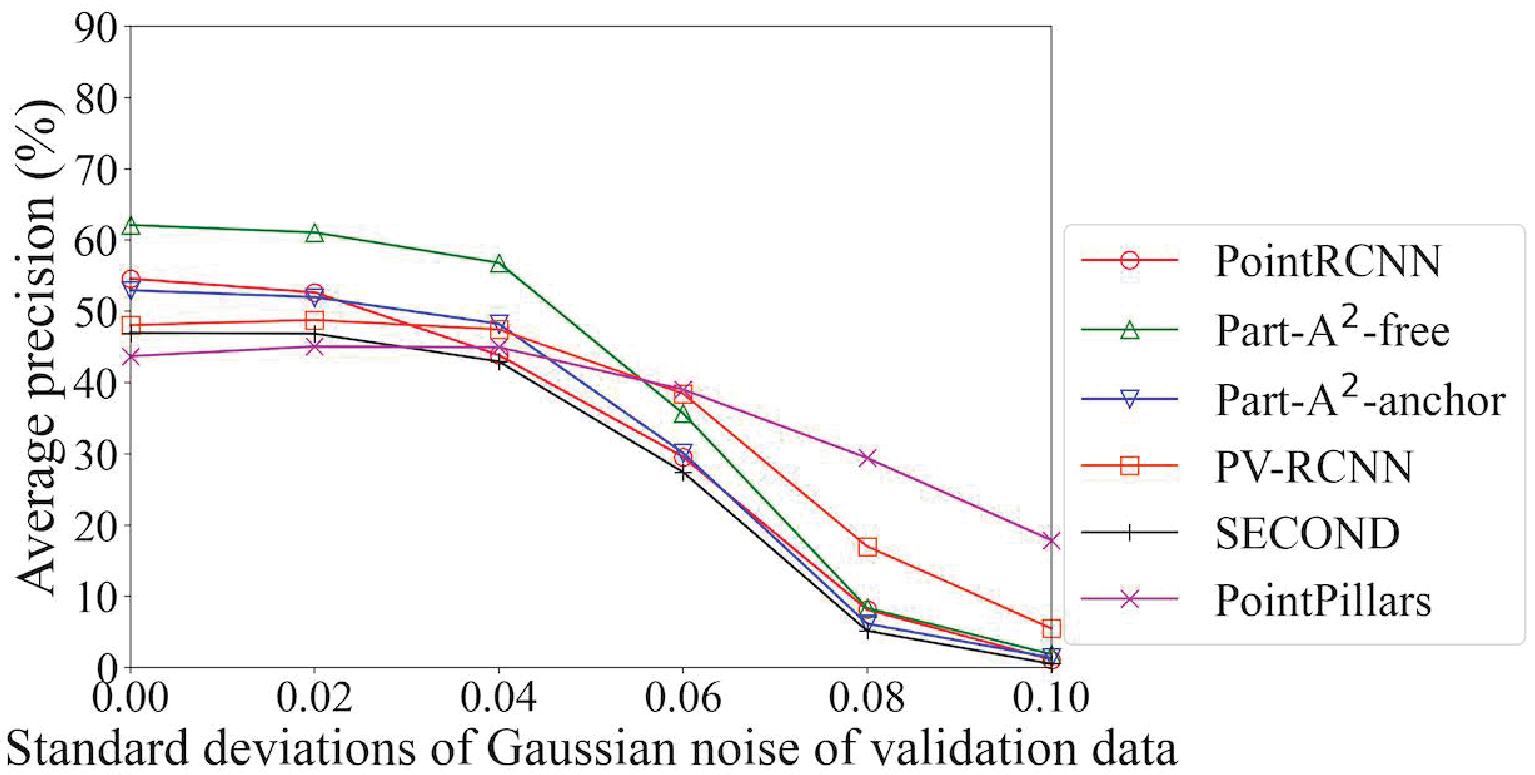}}
	\\
	\subfloat[Cyclist]{\includegraphics[clip, width=\linewidth]{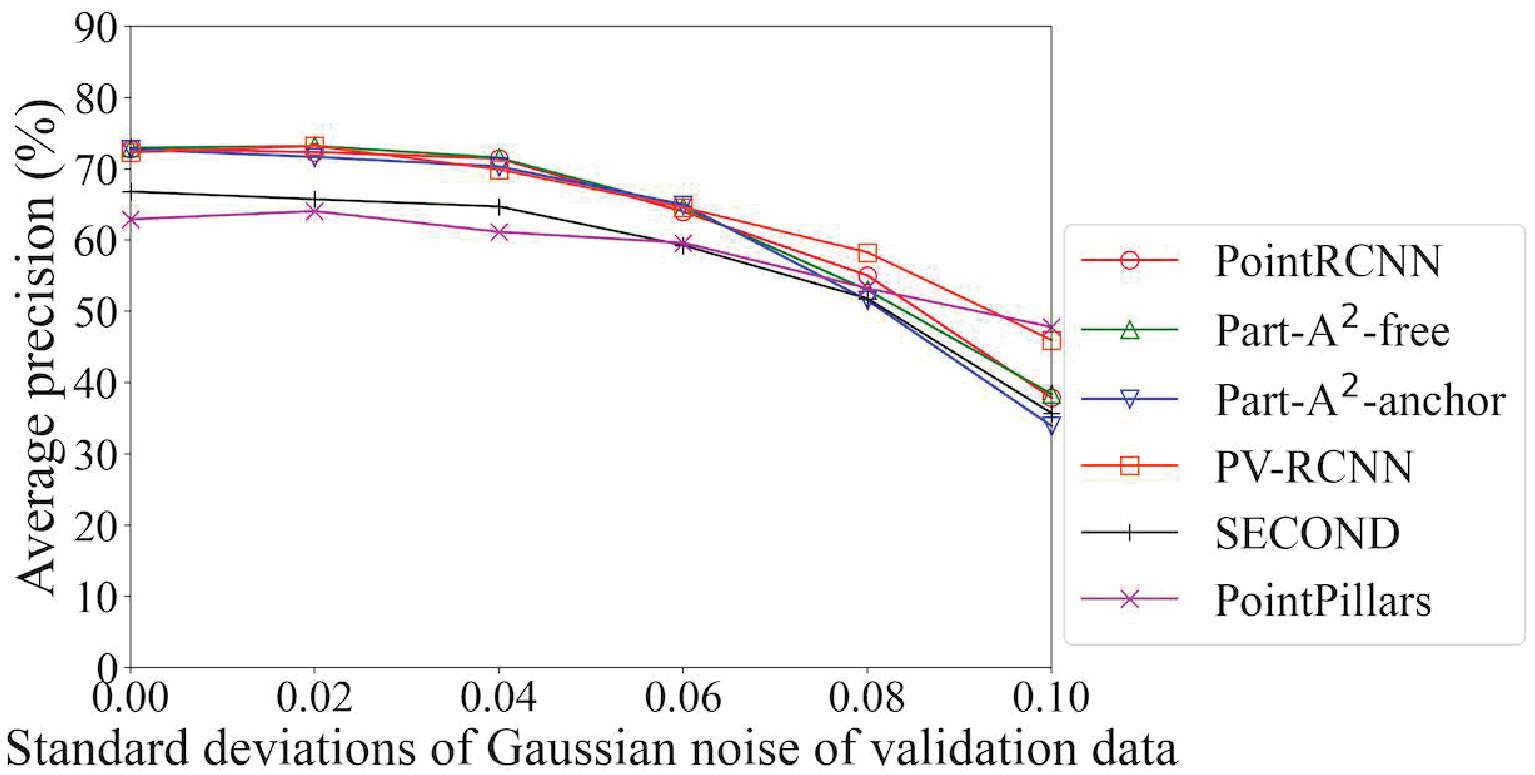}}
	
	\caption{Detection accuracy of each class of Original models for data with noise. The horizontal axis represents the standard deviations of Gaussian noise of the validation data, where 0.00 refers to the original data.}
	\label{fig:Results noise}
\end{figure}

Fig.~\ref{fig:Results noise} shows the detection accuracy of the Original models of each method for data with noise. Among the six methods, only PointPillars adopts the pillar-based process. In light of this, it can be seen that the accuracy of the method that adopts the pillar-based process was higher than others in most cases for data with severe noise, especially for data with noise with standard deviations of 0.08 or higher. For example, the accuracy of car detection for data with noise of standard deviation of 0.10 can differ by up to 32\% (PointPillars with pillar-based process vs. Part-A$ ^2 $-anchor with voxel-based process). However, in addition to the point-based process, the voxel-based process also showed a sharp drop in accuracy due to noise. This is probably because the quantization granularity of the voxel is finer than that of the pillar; the effect of noise was not fully mitigated. This result confirms that the DL method that achieves the highest detection accuracy changes depending on the noise level, and that the selection procedure in the proposed framework can work effectively. It also suggests that a model of the DL methods that adopt the pillar-based process should be selected for data with severe noise.

\subsection{Effect of adding noise to training data} 
\label{sec:Results noisey model}

\begin{figure}[t]
	\centering
	\subfloat[Car]{\includegraphics[clip, width=\linewidth]{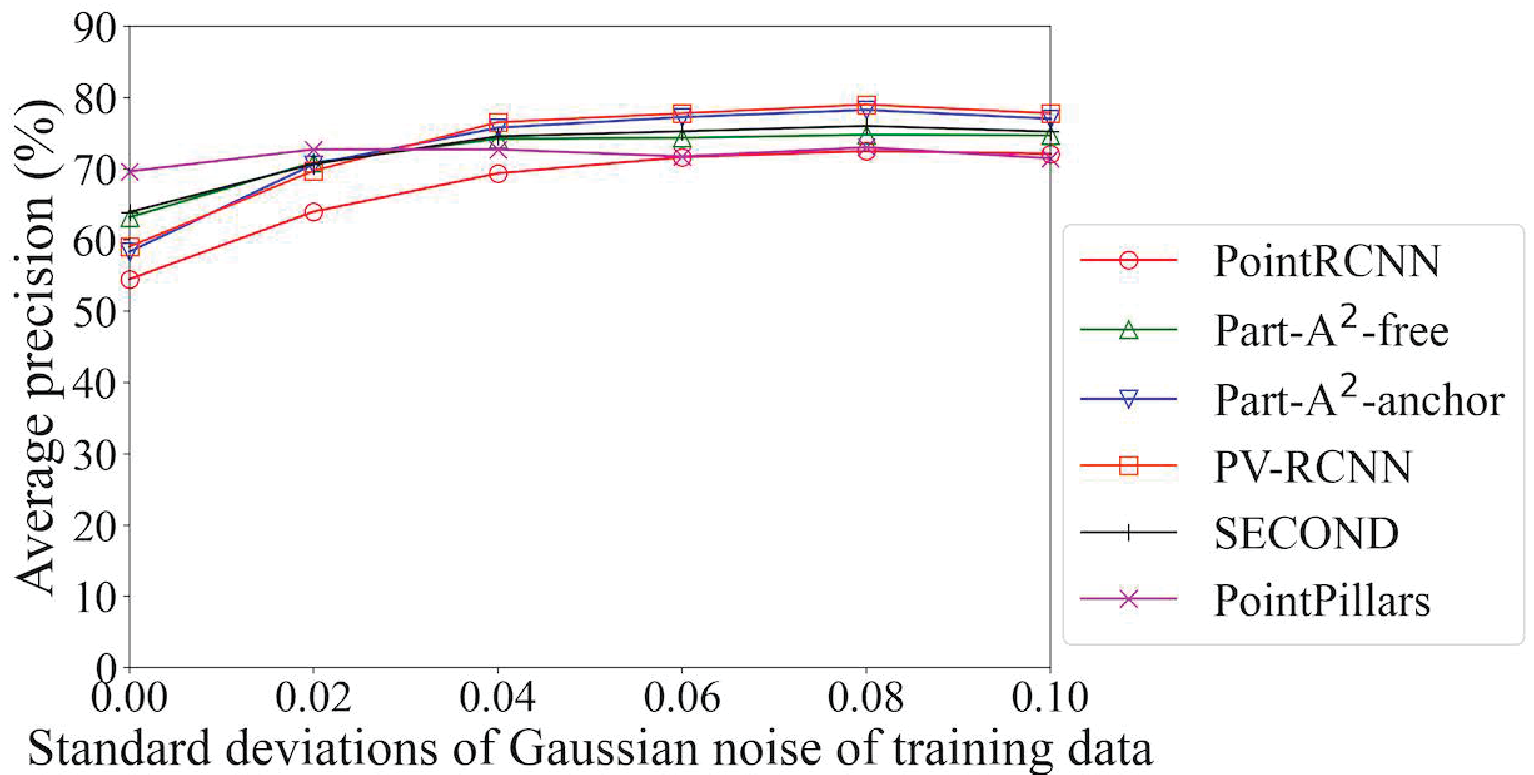}}
	\\
	\subfloat[Pedestrian]{\includegraphics[clip, width=\linewidth]{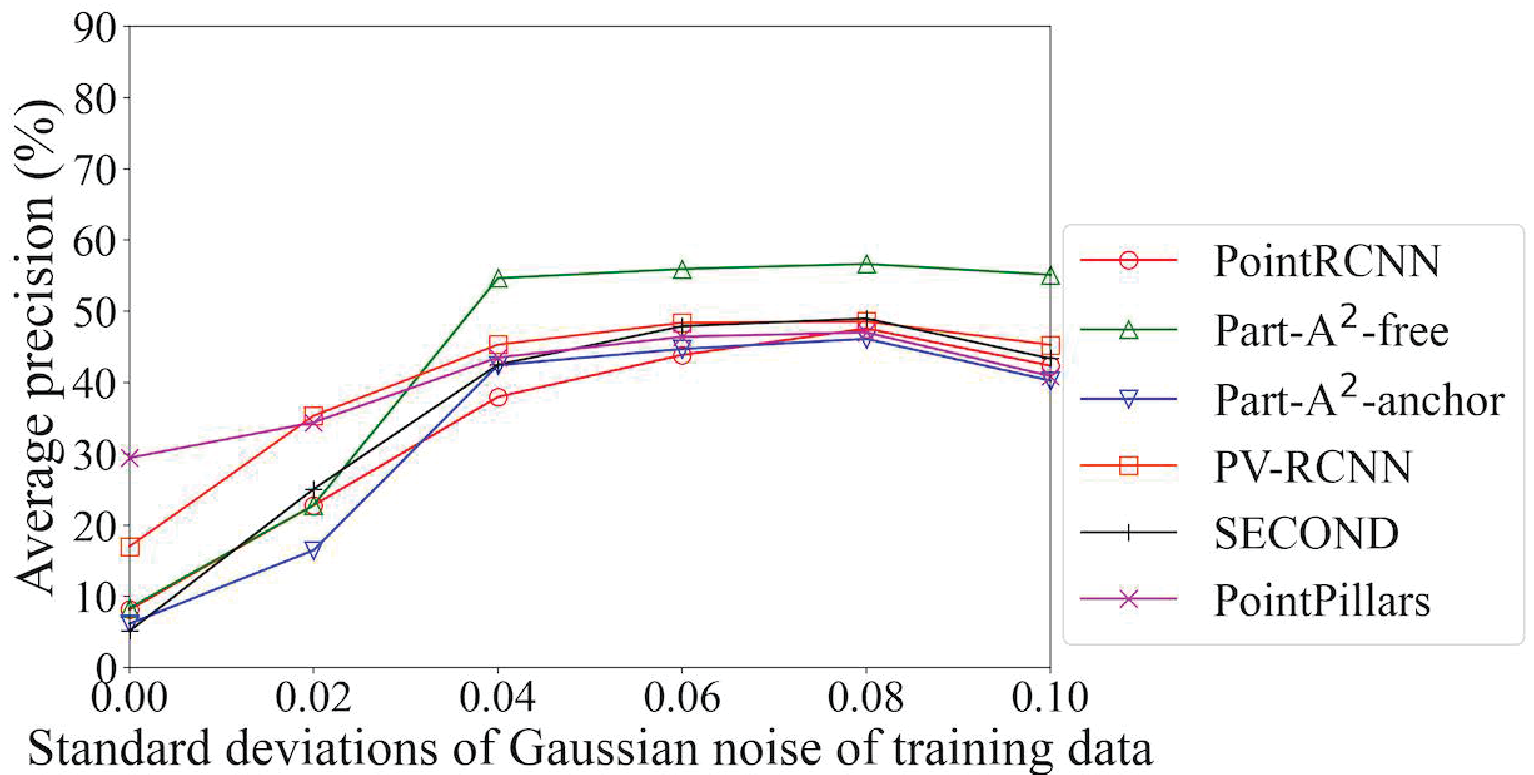}\label{subfig:Results noise ped}}
	\\
	\subfloat[Cyclist]{\includegraphics[clip, width=\linewidth]{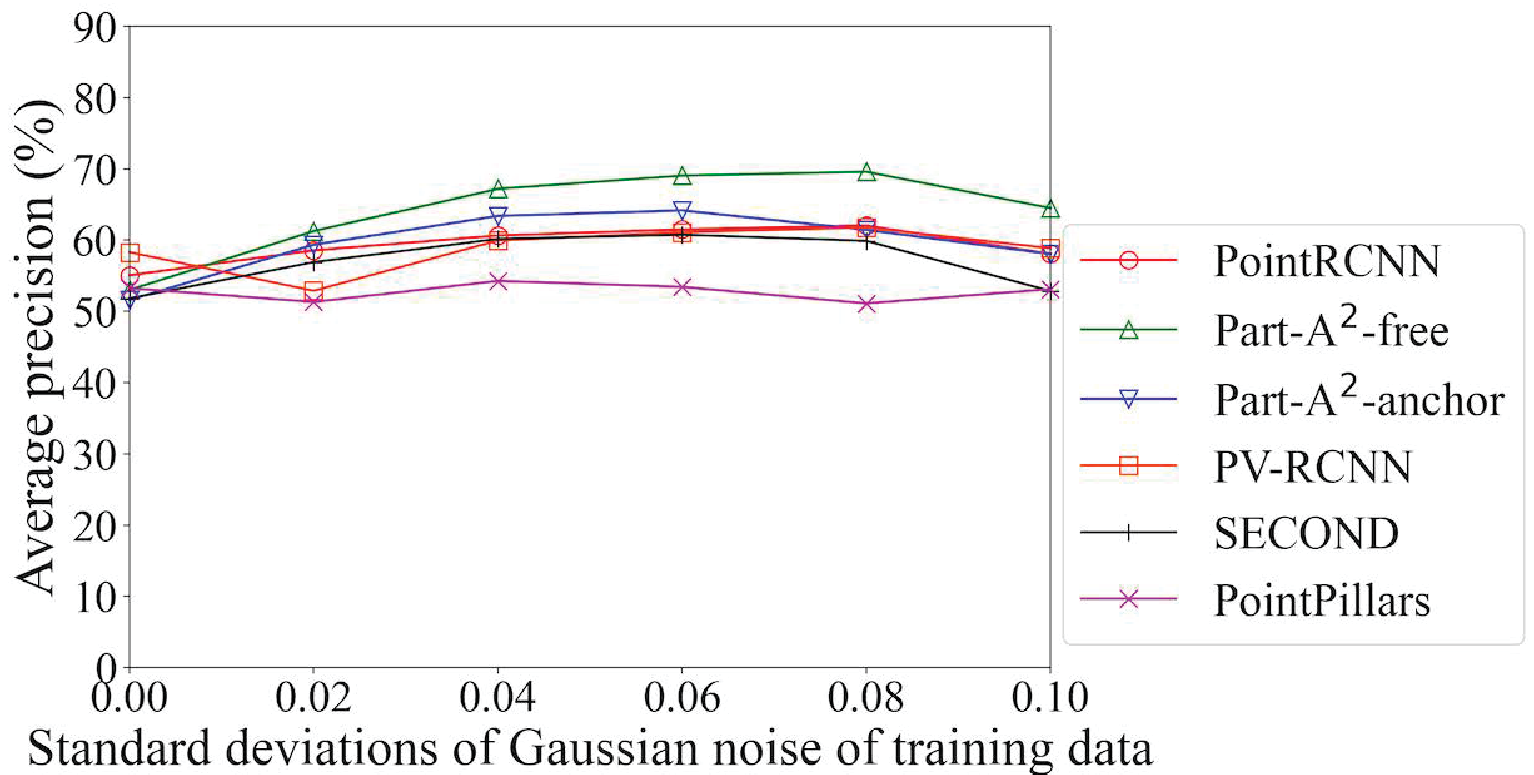}}
	
	\caption{Detection accuracy of Noise models for data with Gaussian noise with standard deviation of 0.08. The horizontal axis represents the standard deviation of the noise of the training data, where 0.00 refers to the Original model.}
	\label{fig:Results noise model}
\end{figure}

\begin{table}[bt]
	\caption{List of models that achieve maximum accuracy for data with Gaussian noise with standard deviation of 0.08. For each method, the model with the maximum accuracy in each class is summarized out of three models (two Noise models and the Original model).}
	\begin{center}
		\renewcommand{\arraystretch}{1.4}
		\begin{tabular}{|c|c|c|c|}
			\hline
			\textbf{DL method} & \textbf{Car} & \textbf{Pedestrian} & \textbf{Cyclist} \\ \hline
			\textbf{PointRCNN} & 0.08 & 0.08 & 0.08 \\ \hline
			\textbf{Part-A$ ^2 $-free} & 0.08 & 0.08 & 0.08 \\ \hline
			\textbf{Part-A$ ^2 $-anchor} & 0.08 & 0.08 & 0.06 \\ \hline
			\textbf{PV-RCNN} & 0.08 & 0.08 & 0.08 \\ \hline
			\textbf{SECOND} & 0.08 & 0.08 & 0.06 \\ \hline
			\textbf{PointPillars} & 0.08 & 0.08 & 0.04 \\ \hline
		\end{tabular}
		\label{tab:Resultus noise model}
	\end{center}
\end{table}

Fig.~\ref{fig:Results noise model} shows the detection accuracy of the Noise models for data with Gaussian noise with the standard deviation of 0.08. Table \ref{tab:Resultus noise model} also summarizes the models with maximum accuracy for each method and each class. In the case of noise, we can see in Table \ref{tab:Resultus noise model} that for almost all items, the models trained on the data with noise close to that of the inference data (standard deviation of 0.08 in this case) achieved the maximum accuracy. The results in Fig.~\ref{fig:Results noise model} show that the improvement in the accuracy of methods that adopt the voxel-based process was larger than those of methods that adopt point-based and pillar-based processes. In particular, PointRCNN, which adopts only the point-based process, had high accuracy in detecting pedestrians and cyclists when the incompleteness was small (Fig.~\ref{fig:Results noise}). These results confirm that the selection procedure in the proposed framework can work effectively. They also suggest that a model of the DL method that adopts the voxel-based process should be selected when using a noise model.

Here, the effect of the proposed framework is reviewed in detail. As in Section~\ref{sec:Results sampling model}, we refer to PointRCNN Original model as the benchmark. For example, in the detection of pedestrians for data with noise shown in Fig.~\ref{subfig:Results noise ped}, our proposed framework selects Part-A$ ^2 $-free Noise 0.08 model according to the flowchart in Fig.~\ref{fig:Model selection}. In this case, the proposed framework improves the average precision by 48\% compared to the benchmark.

\subsubsection{Consideration of model selection by score} 
\label{sec: Results score}

\begin{figure}[t]
	\centering
	\subfloat[Car (PV-RCNN)]{\includegraphics[clip, width=0.9\linewidth]{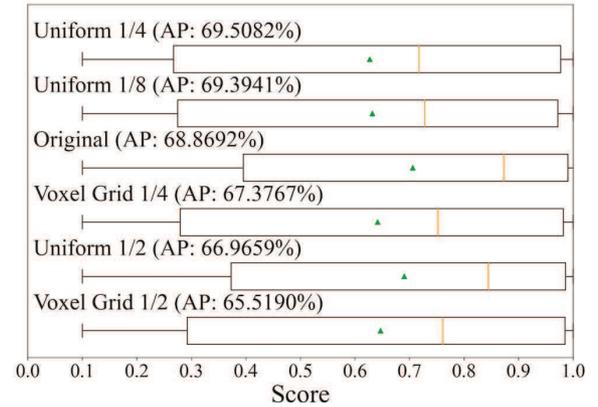}}
	\\
	\vspace{-12pt}
	\subfloat[Pedestrian (PointRCNN)]{\includegraphics[clip, width=0.9\linewidth]{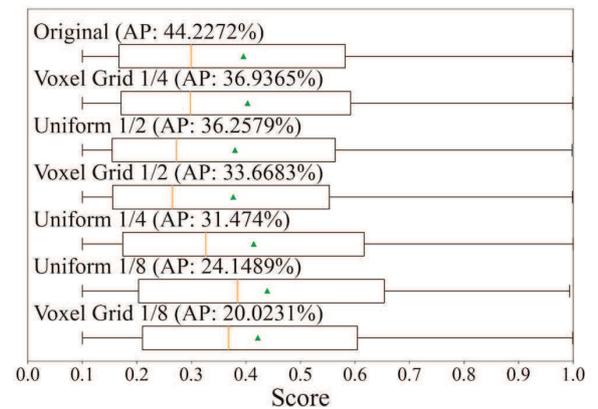}}
	\\
	\vspace{-12pt}
	\subfloat[Cyclist (PointRCNN)]{\includegraphics[clip, width=0.9\linewidth]{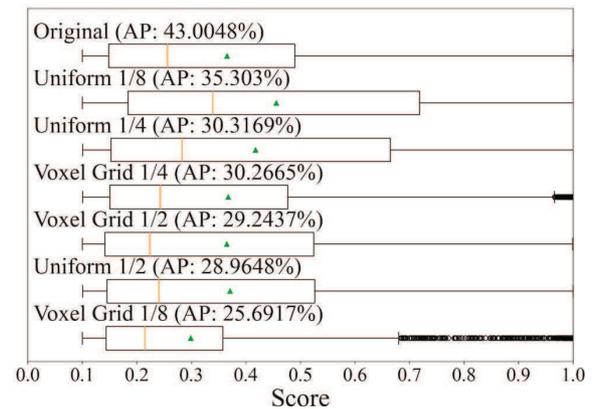}}
	
	\caption{Relationship between score distributions of detected boxes and detection accuracies for data with normalized number of points of 0.25. AP denotes the average precision. Models are listed from top to bottom in order of detection accuracy.}
	\label{fig:Results score}
\end{figure}

We next consider whether the best model could be preselected in accordance with the scores obtained when training models or not.
Fig.~\ref{fig:Results score} shows the relationship between the score distributions of each box detected by each model and the detection accuracies for data with the normalized number of points of 0.25. The score distributions are represented by box plots, and the results of the sampling models of PV-RCNN and PointRCNN, which have the model with the maximum accuracy in each class, are given as examples. Looking at the median (orange line) and mean (green triangle), we can see that the score distribution of the models with high detection accuracy is not necessarily higher than that of the others. This is a mixed situation in which a certain number of objects could be detected without confidence, a certain number of objects could be detected with confidence, and there were a lot of false negatives. This result indicates that the model selection based on the score distribution of the output cannot properly select a model with high accuracy for the data with low density of points. In contrast, the model selection based on the features in our framework can properly select a model with high accuracy, as described in Section~\ref{sec:Results sampling model}.

\section{Outdoor experiment examples} 
\label{sec:Outdoor}

\begin{figure}[t]
	\centerline{\includegraphics[width=0.9\linewidth]{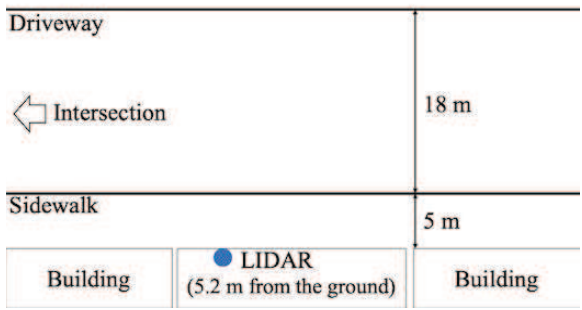}}
	\caption{Layout of road near intersection used for outdoor experiment. Blue circle indicates the location of the LIDAR unit.}
	\label{fig:Outdoor place}
\end{figure}

\begin{figure}[t]
	\centering
	
	\subfloat[2021 2/5 16:39:27]{\includegraphics[clip, width=0.7\linewidth]{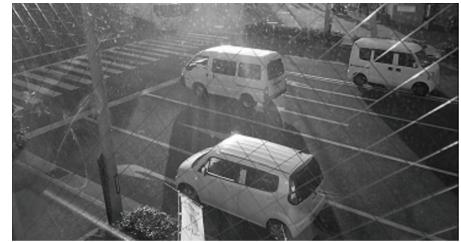}\label{subfig:Outdoor camera 1}}
	\\
	\vspace{-5pt}
	\subfloat[2021 2/5 16:40:56]{\includegraphics[clip, width=0.7\linewidth]{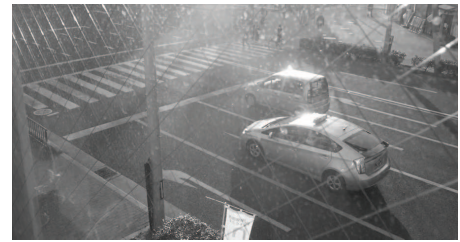}\label{subfig:Outdoor camera 2}}
	\\
	\vspace{-5pt}
	\subfloat[2021 2/5 16:41:56]{\includegraphics[clip, width=0.7\linewidth]{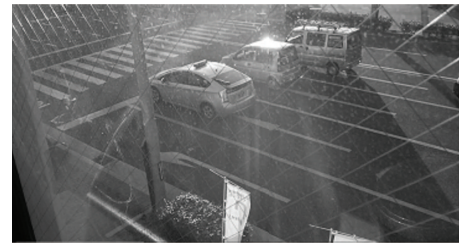}\label{subfig:Outdoor camera 3}}
	\\
	\vspace{-5pt}
	\subfloat[2021 2/5 16:43:27]{\includegraphics[clip, width=0.7\linewidth]{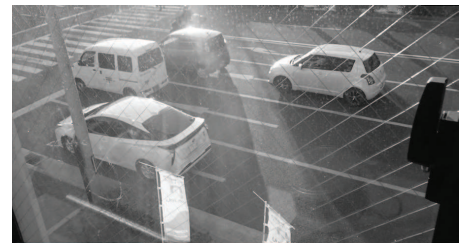}\label{subfig:Outdoor camera 4}}
	
	\caption{Pictures of intersection. They have been changed to grayscale and reduced in resolution to ensure privacy. Below each picture is the date and time it was taken.}
	\label{fig:Outdoor camera}
\end{figure}

\begin{figure}[t]
	\subfloat[PV-RCNN Original model]{
		\includegraphics[clip, width=0.45\linewidth]{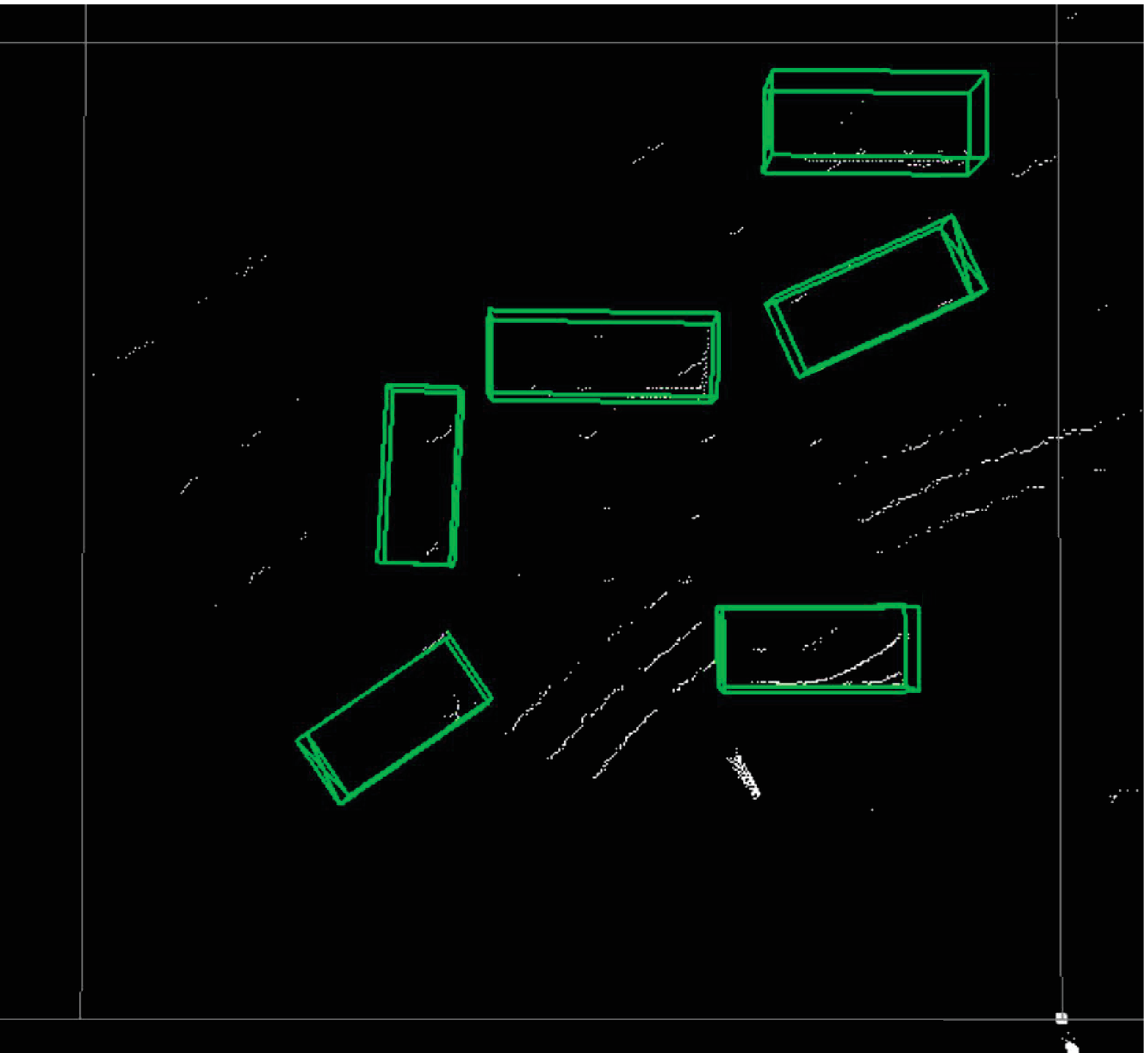}} \quad
	\subfloat[PV-RCNN Uniform 1/8 model]{
		\includegraphics[clip, width=0.45\linewidth]{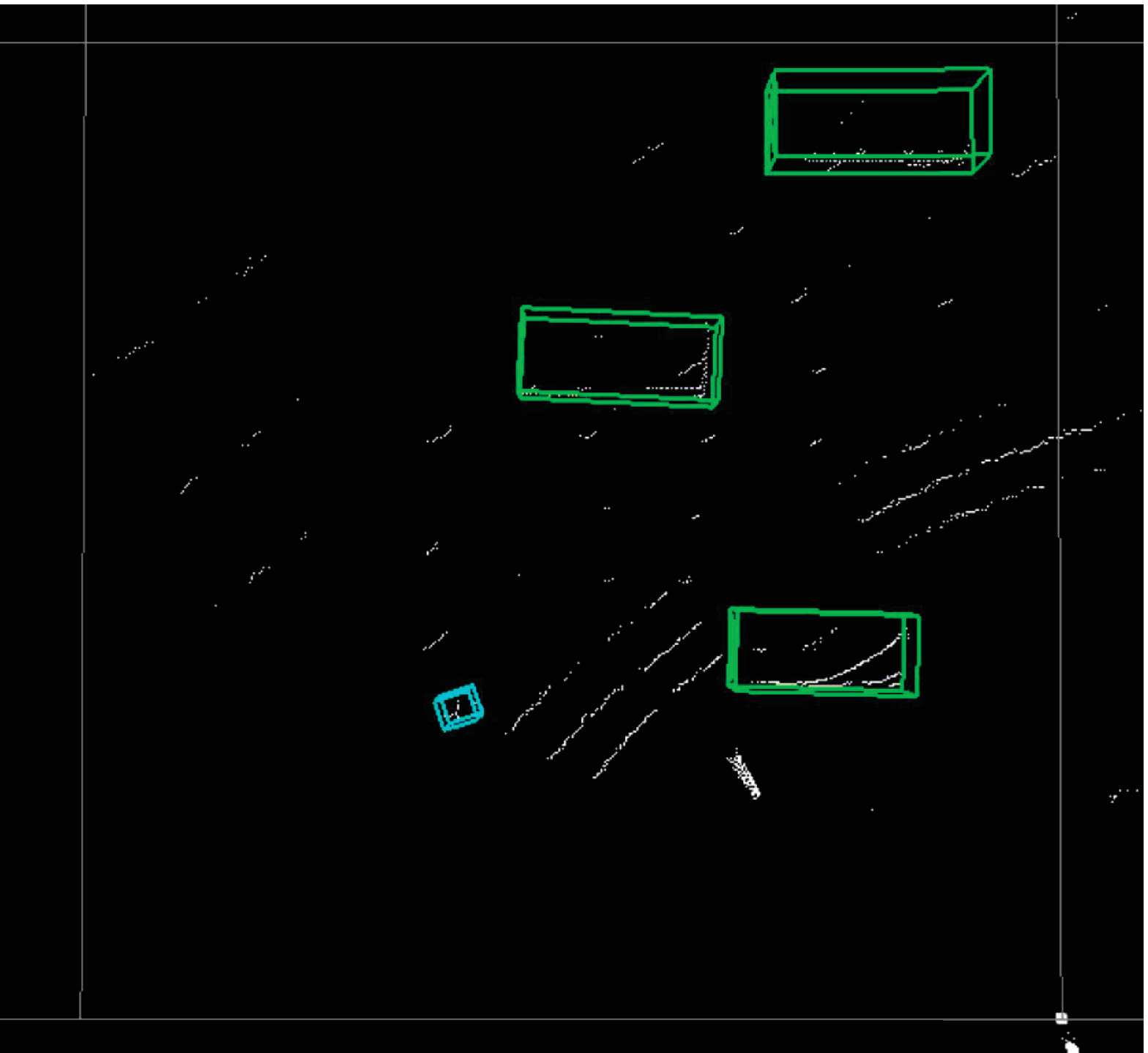}}
	\caption{Results of object detection for point cloud data corresponding to Fig.~\ref{subfig:Outdoor camera 1}. Green boxes represent the car's bounding boxes.}
	\label{fig:Outdoor result 1}
\end{figure}

\begin{figure}[t]
	\vspace{-10pt}
	\subfloat[PV-RCNN Original model]{
		\includegraphics[clip, width=0.45\linewidth]{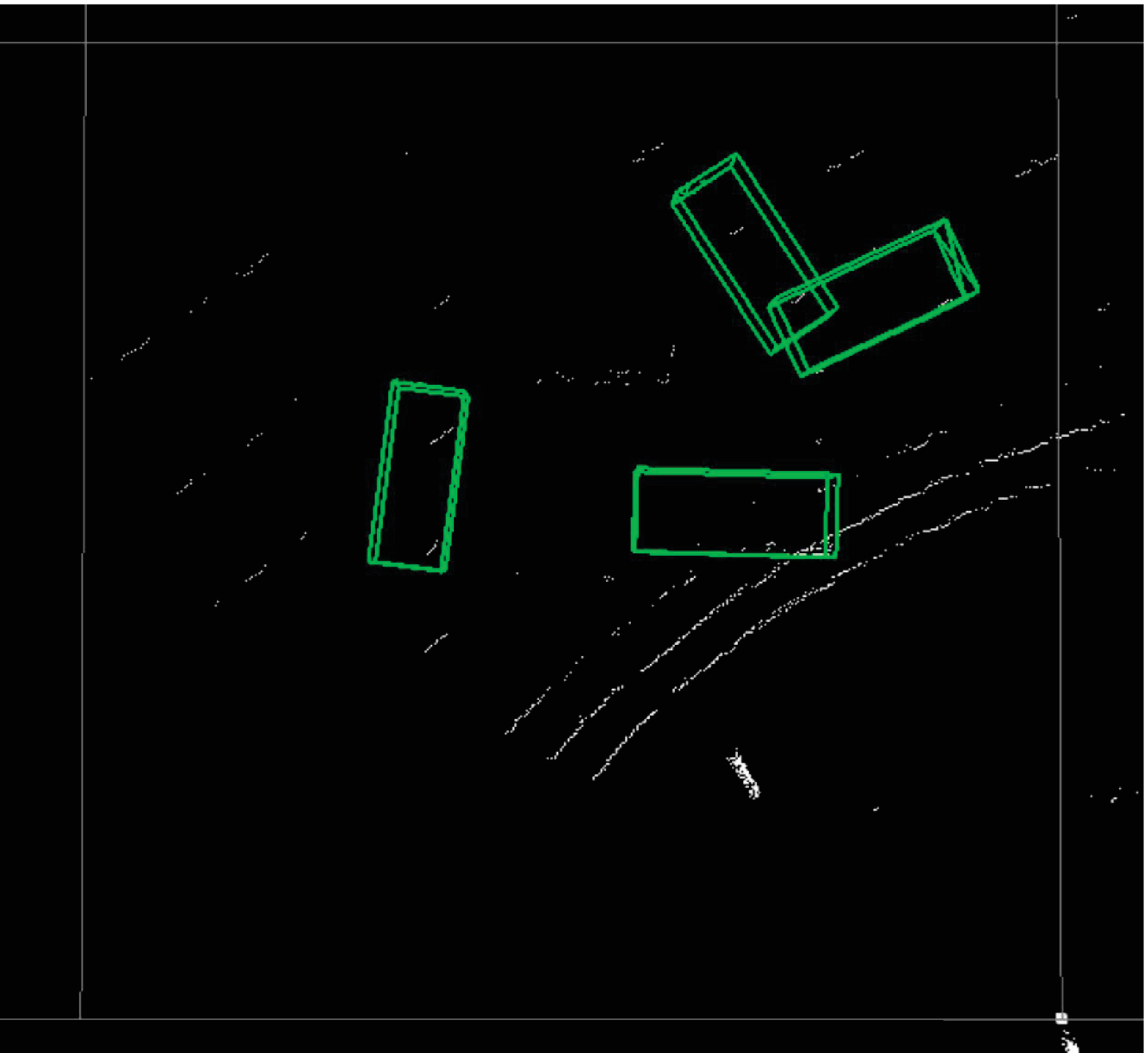}} \quad
	\subfloat[PV-RCNN Uniform 1/8 model]{
		\includegraphics[clip, width=0.45\linewidth]{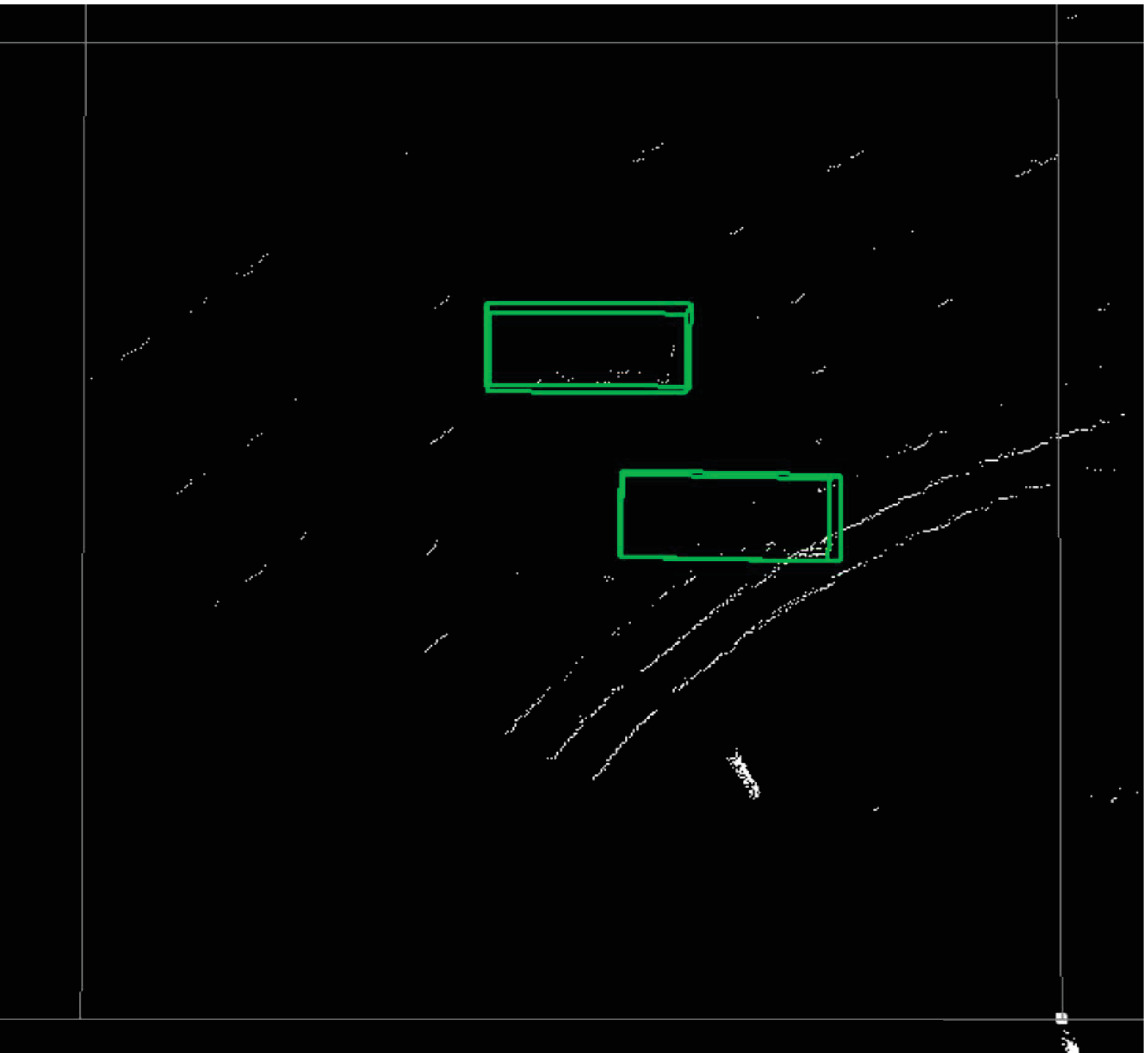}}
	\caption{Results of object detection for point cloud data corresponding to Fig.~\ref{subfig:Outdoor camera 2}. Green boxes represent the car's bounding boxes.}
	\label{fig:Outdoor result 2}
\end{figure}

\begin{figure}[t]
	\vspace{-10pt}
	\subfloat[PV-RCNN Original model]{
		\includegraphics[clip, width=0.45\linewidth]{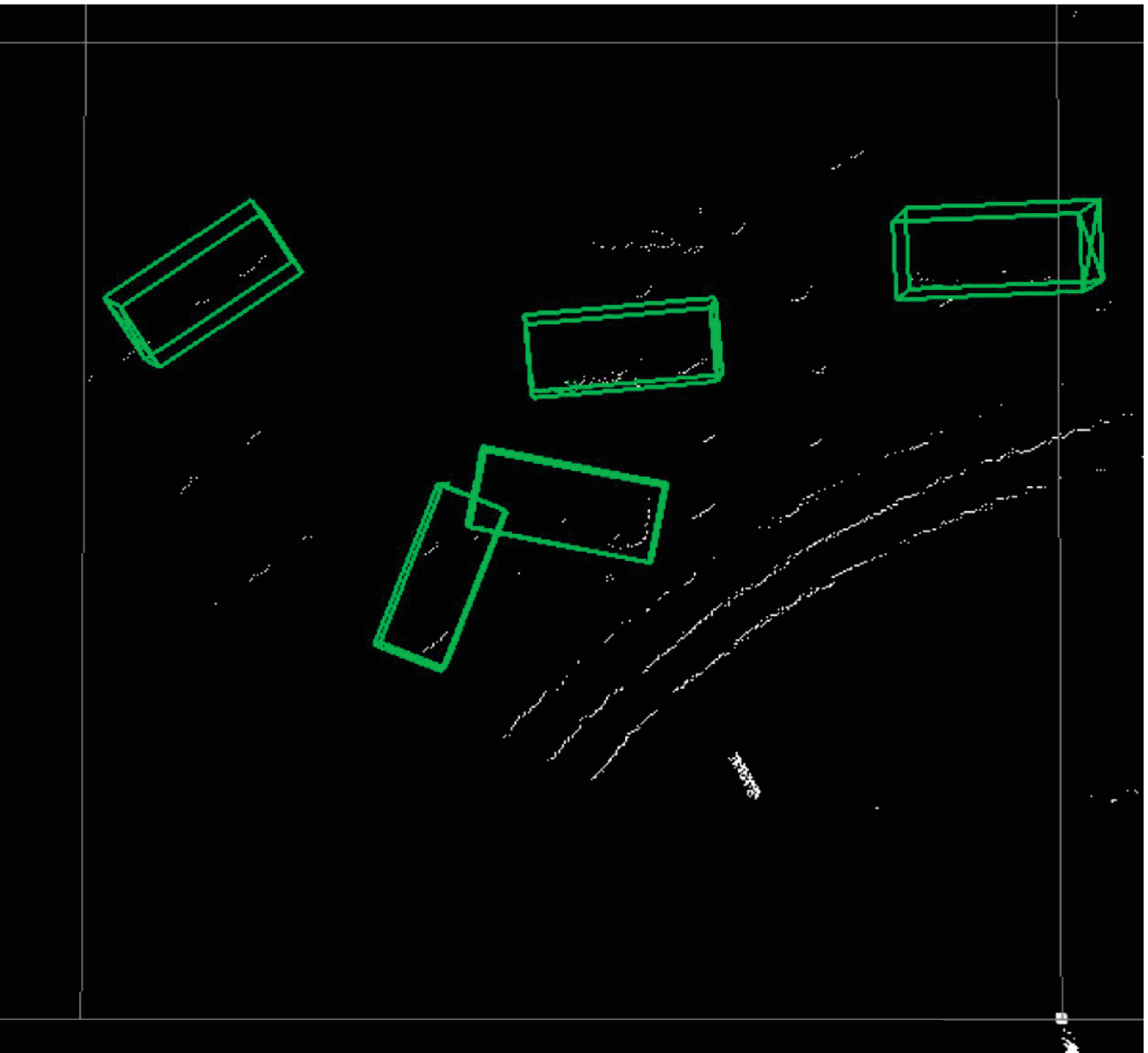}} \quad
	\subfloat[PV-RCNN Uniform 1/8 model]{
		\includegraphics[clip, width=0.45\linewidth]{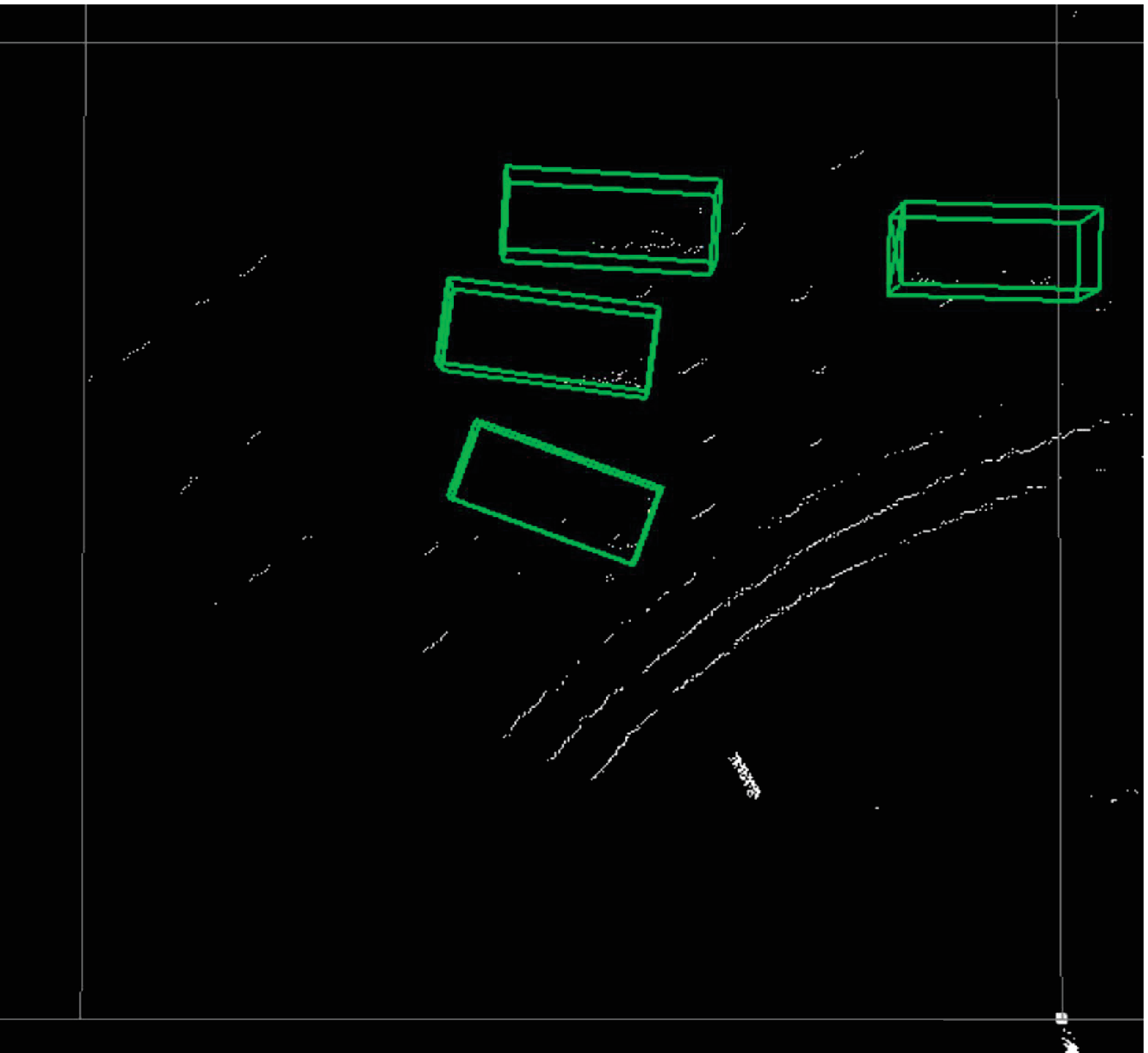}}
	\caption{Results of object detection for point cloud data corresponding to Fig.~\ref{subfig:Outdoor camera 3}. Green boxes represent the car's bounding boxes.}
	\label{fig:Outdoor result 3}
\end{figure}

\begin{figure}[t]
	\vspace{-10pt}
	\subfloat[PV-RCNN Original model]{
		\includegraphics[clip,width=0.45\linewidth]{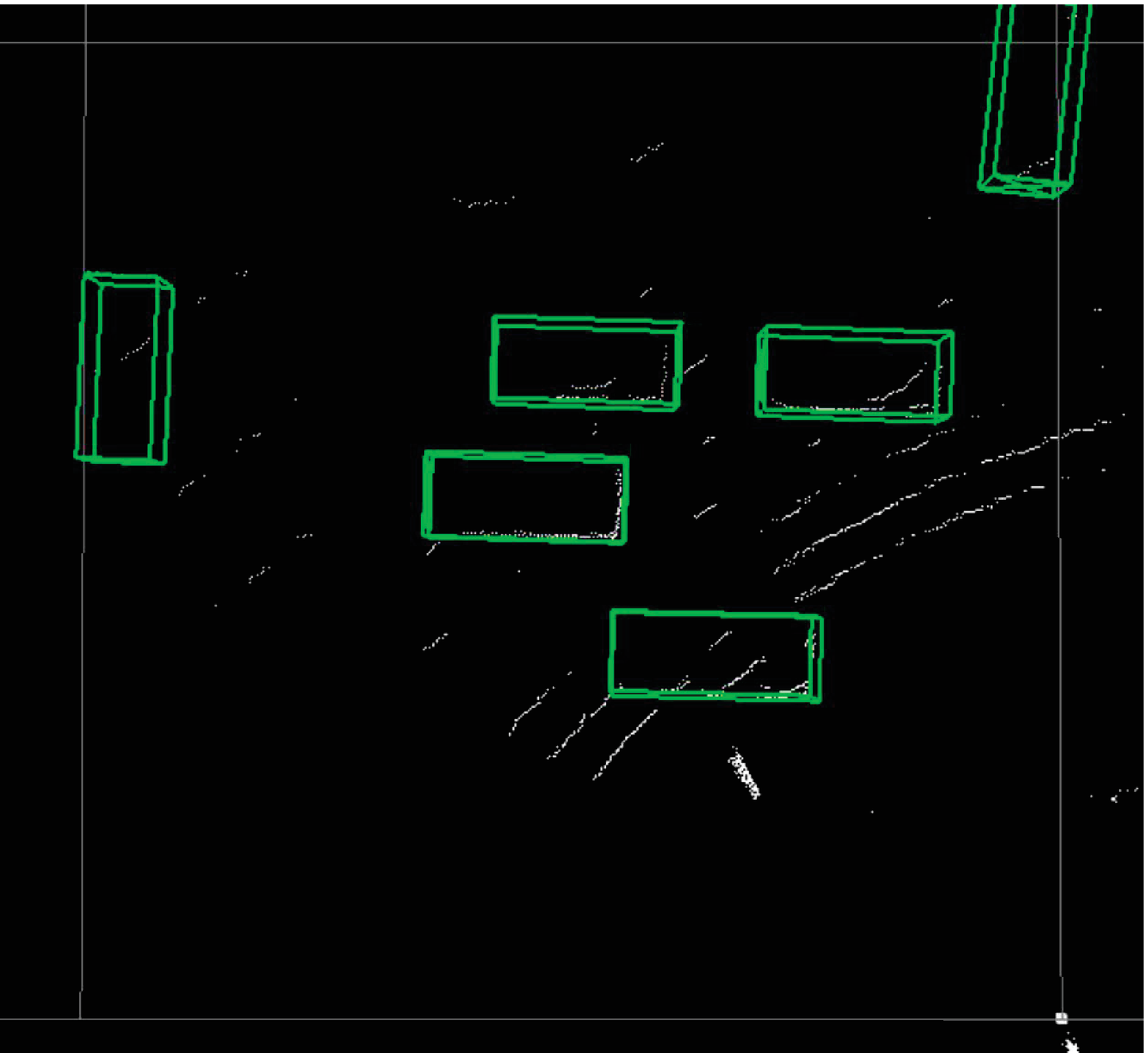}} \quad
	\subfloat[PV-RCNN Uniform 1/8 model]{
		\includegraphics[clip, width=0.45\linewidth]{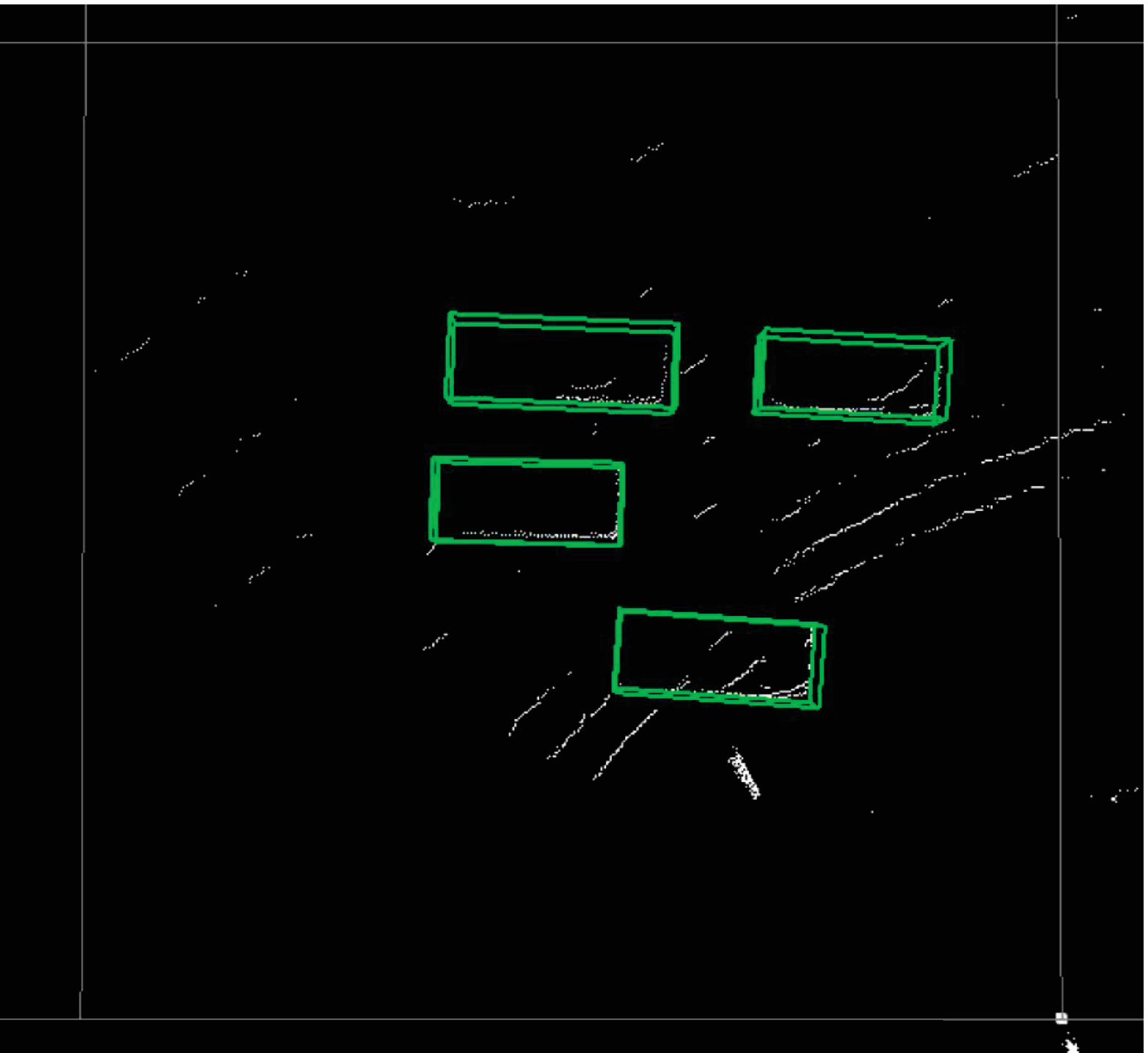}}
	\caption{Results of object detection for point cloud data corresponding to Fig.~\ref{subfig:Outdoor camera 4}. Green boxes represent the car's bounding boxes.}
	\label{fig:Outdoor result 4}
\end{figure}

Finally, we present examples from the outdoor experiment using data measured in an actual outdoor environment. Fig.~\ref{fig:Outdoor place} shows the layout of the road near the intersection (Hyakumanben Intersection, Kyoto City, Kyoto Prefecture, Japan) used for this experiment, which includes the setup position of the LIDAR unit we used (Velodyne VLP-16 LIDAR). Fig.~\ref{fig:Outdoor camera} shows pictures for checking answers that show the intersection when the point cloud data were actually acquired. This outdoor experiment focused on car detection and evaluated the performance of the Uniform model for data with low density of points. In accordance with the results in Section~\ref{sec:Results density}, the method we used was PV-RCNN, which adopts an anchor-based strategy and point-based process. In addition, since the normalized number of points in the acquired point cloud data was about 0.080, the Uniform 1/8 model trained on the data with the closest normalized number of points was compared to the Original model. Figs.~\ref{fig:Outdoor result 1}--\ref{fig:Outdoor result 4} show the visualization of the detection results for point cloud data corresponding to each picture in Fig.~\ref{fig:Outdoor camera}. These figures show the results looking down from above. In Fig.~\ref{fig:Outdoor result 1}, we can see that both models were able to detect all cars in the picture, though the PV-RCNN Original model had three false positives. In Fig.~\ref{fig:Outdoor result 2}, we can see that the PV-RCNN Uniform 1/8 model was able to correctly detect the two cars in the picture, while the PV-RCNN Original model was only able to detect the one in front of it and had three false positives. In Fig.~\ref{fig:Outdoor result 3}, we can also see that the PV-RCNN Uniform 1/8 model was able to detect all cars in the picture, while the PV-RCNN Original model was unable to properly detect the car on the furthest side and had two false positives. In Fig.~\ref{fig:Outdoor result 4}, we can see that both models were able to detect all cars in the picture, though the PV-RCNN Original model had two false positives. These four cases demonstrate that the Uniform model had an advantage in terms of detection accuracy. Therefore, the results of this outdoor experiment are useful for confirming the effectiveness of the Uniform model for data that have dropped in density of points due to differences of performance between 3D image sensors.

\section{Conclusion} 
\label{sec:Conclusion}
In this paper, we proposed a feature-based model selection framework to deal with various features of point cloud data acquired by 3D image sensors. The proposed framework uses the broad features of DL methods to select a suitable model in accordance with the features of the point cloud data. The proposed framework uses training data with pseudo incompleteness to reduce the difference between the training and inference data. In the evaluation, we compared multiple models obtained from multiple DL methods with sampling and adding noise to show the effectiveness of the feature-based model selection in the proposed framework. The evaluation results verified that it effectively works to select a suitable DL model in accordance with the features of three factors, i.e., object class, density of points, and noise level. We also demonstrated through an outdoor experiment that the DL model constructed using the training data with a similar feature to the data in the real measurement environment works better than the DL model constructed using the original data.

\section*{Acknowledgments}
The evaluation results were partly obtained from research commissioned by the National Institute of Information and Communications Technology (NICT), Japan.

\bibliographystyle{ieicetr}
\bibliography{references}

\profile{Kairi Tokuda}{is currently pursuing his master's degree at the Graduate School of Informatics, Kyoto University, Kyoto, Japan. He received his B.E. from Kyoto University, Kyoto, Japan, in 2021. His research interests include smart monitoring, machine learning, and virtual networks.}

\clearpage

\profile{Ryoichi SHINKUMA}{received B.E., M.E., and Ph.D. degrees in communications engineering from Osaka University in 2000, 2001, and 2003, respectively. He joined the Graduate School of Informatics, Kyoto University and worked there as an assistant professor from 2003 to 2011 and as an associate professor from 2011 to 2021. He was a visiting scholar at the Wireless Information Network Laboratory, Rutgers University from 2008 to 2009. In 2021, he joined the Faculty of Engineering, Shibaura Institute of Technology as a professor. His main research interest is cooperation in heterogeneous networks. He received the young researchers' award, the best tutorial paper award of the Communications Society, the best paper award from the IEICE in 2006, 2019, and 2022, respectively. He also received the Young Scientist Award from Ericsson Japan in 2007 and the TELECOM System Technology Award from the Telecommunications Advancement Foundation in 2016. He is a fellow of the IEICE and a senior member of the IEEE.}

\profile{Takehiro SATO}{received his B.E., M.E., and Ph.D. degrees in engineering from Keio University in 2010, 2011, and 2016, respectively. He is currently an associate professor at the Graduate School of Informatics at Kyoto University. His research interests include communication protocols and network architecture for next-generation optical networks. From 2011 to 2012, he was a research assistant in the Keio University Global COE Program, ``High-level Global Cooperation for Leading-edge Platform on Access Spaces,'' established by the Ministry of Education, Culture, Sports, Science and Technology of Japan. From 2012 to 2015, he was a research fellow with the Japan Society for the Promotion of Science. From 2016 to 2017, he was a research associate at the Graduate School of Science and Technology at Keio University. He is a member of the IEEE and IEICE.}

\profile{Eiji OKI}{received B.E. and M.E. degrees in instrumentation engineering and a Ph.D. in electrical engineering from Keio University, Yokohama, Japan, in 1991, 1993, and 1999. He was with Nippon Telegraph and Telephone Corporation (NTT) Laboratories, Tokyo, from 1993 to 2008, and the University of Electro-Communications, Tokyo, from 2008 to 2017. From 2000 to 2001, he was a Visiting Scholar at Polytechnic University, Brooklyn, New York. In 2017, he joined Kyoto University, Japan, where he is currently a Professor. His research interests include routing, switching, protocols, optimization, and traffic engineering in communication and information networks. He is a Fellow of the IEEE and IEICE.}

\end{document}